\documentclass[lettersize,journal]{IEEEtran}

\usepackage[utf8]{inputenc} 
\usepackage[T1]{fontenc}    
\usepackage{hyperref}       
\usepackage{booktabs}       
\usepackage{amsfonts}       
\usepackage{nicefrac}       
\usepackage{microtype}      
\usepackage{xcolor}         

\usepackage{cite}
\usepackage{amsmath}
\usepackage{amssymb}
\usepackage{algorithm}
\usepackage{algpseudocode}
\usepackage{acronym}
\usepackage{bm}
\usepackage{enumitem}
\usepackage{xspace}
\usepackage[pdftex]{graphicx}
\usepackage{amsthm}

\acrodef{fl}[FL]{federated learning}
\acrodef{sgd}[SGD]{stochastic gradient descent}
\acrodef{fa}[FedAvg]{federated averaging}

\newcommand{\Asaf}[1]{\textcolor{blue}
{\footnotesize{\textsf {[Asaf: #1]}}}}
\newcommand{\ale}[1]{\textcolor{red}{[ #1 -- Alejandro ] \normalsize}}

\newcommand{\changed}[1]{\textcolor{red}{#1}}

\algdef{SE}[SUBALG]{Indent}{EndIndent}{}{\algorithmicend\ }%
\algtext*{Indent}
\algtext*{EndIndent}

\newcommand{\cD}{\mathcal{D}}
\newcommand{\cU}{\mathcal{U}}

\def\vw{{\bm{w}}}

\newcommand{\R}{\mathbb{R}}

\newtheorem{lemma}{Lemma}
\newtheorem{theorem}{Theorem}
\newtheorem{problem}{Problem}

\usepackage{amsmath,amsfonts}
\usepackage{array}
\usepackage[caption=false,font=normalsize,labelfont=sf,textfont=sf]{subfig}
\usepackage{comment}
\usepackage{textcomp}
\usepackage{stfloats}
\usepackage{url}
\usepackage{verbatim}
\usepackage{graphicx}
\def\BibTeX{{\rm B\kern-.05em{\sc i\kern-.025em b}\kern-.08em
    T\kern-.1667em\lower.7ex\hbox{E}\kern-.125emX}}
\usepackage{balance}

\usepackage[all=normal,paragraphs=tight,floats=normal,mathspacing=normal,wordspacing=tight,charwidths=tight,mathdisplays=normal,leading=normal]{savetrees}

\begin{document}
\title{Adaptive Deadline and Batch Layered\\ Synchronized Federated Learning}
\author{
\IEEEauthorblockN{Asaf Goren\IEEEauthorrefmark{1}, Natalie Lang\IEEEauthorrefmark{2}, Nir Shlezinger\IEEEauthorrefmark{2}, Alejandro Cohen\IEEEauthorrefmark{1}}\\
\IEEEauthorblockA{\IEEEauthorrefmark{1}Faculty of ECE, Technion, Haifa, Israel, Emails: asaf.goren@campus.technion.ac.il and alecohen@technion.ac.il}\\
\IEEEauthorblockA{\IEEEauthorrefmark{2}School of ECE, Ben-Gurion University, Beer-Sheva, Israel, Emails:  langn@post.bgu.ac.il and  nirshl@bgu.ac.il\vspace{-0.6cm}}
}


\maketitle

\begin{abstract}
Federated learning (FL) enables collaborative model training across distributed edge devices while preserving data privacy, and typically operates in a round-based synchronous manner.
However, synchronous FL suffers from latency bottlenecks due to device heterogeneity, where slower clients (stragglers) delay or degrade global updates.
Prior solutions, such as fixed deadlines, client selection, and layer-wise partial aggregation, alleviate the effect of stragglers, but treat round timing and local workload as static parameters, limiting their effectiveness under strict time constraints.
We propose \emph{Adaptive Deadline and Batch Layered Federated Learning} (ADEL-FL), a novel framework that jointly optimizes per-round deadlines and user-specific batch sizes for layer-wise aggregation.
Our approach formulates a constrained optimization problem minimizing the expected $\ell_2$ distance to the global optimum under total training time and global rounds.
We provide a convergence analysis under exponential compute models and prove that ADEL-FL yields unbiased updates with bounded variance.
Extensive experiments demonstrate that ADEL-FL outperforms alternative methods in both convergence rate and final accuracy under heterogeneous conditions.
\end{abstract}


\section{Introduction} \label{sec:introduction}
\IEEEPARstart{F}{ederated} learning (FL) is a decentralized machine learning paradigm where multiple clients collaboratively train a shared model under the coordination of a central server, without exchanging their raw data \cite{mcmahan2017communication}.
This approach enables learning from widespread data while preserving  privacy and reducing communication by only transmitting model updates.
FL has opened opportunities to leverage data from edge devices (smartphones, IoT, etc.) for training models while keeping sensitive information local \cite{zhou2021survey,abdulrahman2020survey}.

FL typically operates as a round based iterative procedure~\cite{gafni2021federated}. In each round, the central server transmits the current global model to a set of participating clients.
Each client then uses its local dataset and computational resources to compute an update to the model via, e.g., stochastic gradient descent
(SGD)~\cite{stich2018local}.
Following local computation, these model updates are transmitted back to the server, which aggregates them to form the new global model.
In the standard federated averaging (FedAvg) protocol \cite{mcmahan2017communication}, the server must wait for all (or most) clients to send updates before proceeding, so a slow client (called a {\em straggler}) can become a bottleneck.
This operation combined with the variability in device speed and availability can lead to significant latency and  degrade the model if certain clients rarely contribute due to repeated timeouts.

In practical federated systems, client delays in training arise from two main sources: $(i)$ the time clients spend computing local updates;
and $(ii)$ the time required to communicate these updates to the server \cite{yang2020delay, asad2023limitations}. The communication delay can be partially mitigated by techniques such as sparsification \cite{han2020adaptive,aji2017sparse, alistarh2018convergence,shi2019topksparsification} and compression \cite{wen2017terngrad, yujun2018deepgradientcompression,shlezinger2021UVeQFed,lang2025olala} of the model updates before transmission.
The computational delay often remains a dominant bottleneck in synchronous FL, as empirical distributed learning studies showed that some workers are often significantly slower than the median worker under typical conditions \cite{tandon2017gradient}.

A variety of approaches have been proposed to mitigate straggler impact in distributed learning and FL.
A simple yet common strategy in synchronous SGD is to impose a per-round deadline: the server aggregates whatever updates have arrived by a cutoff time $T$ and ignores late submissions \cite{chen2016revisiting}.
This {\em drop-stragglers} approach has been shown to converge at the same asymptotic rate as full participation \cite{li2019convergence}.
However, discarding stragglers outright can waste useful computation and potentially bias the learned model (since slower devices might consistently be left out).
An alternative approach alters FL to be asynchronous~\cite{xu2023asynchronous}, avoiding waiting for stragglers by incorporating each update as it arrives, at the cost of some updates being stale \cite{xie2020asynchronous}.
Stale updates, i.e., gradients computed on an older version of the global model, can introduce gradient mismatch and slow down or even destabilize convergence \cite{dai2018toward}.
To reduce the impact of stale updates, staleness-aware aggregation methods were proposed, such as buffering or reweighting updates based on arrival time \cite{nguyen2022federated}.
Still, asynchronous FL typically requires the number of slow users to be small for stable learning~\cite{pfeiffer2023federated}, and is thus limited when learning under tight latency constraints. It was proposed in~\cite{ma2021fedsa, zhang2023semi}  to combine asynchronous FL with synchronous operation by synchronizing local stale models while performing global updating asynchronously.
However, such semi-asynchronous FL, there may still be high variation between participating users in their local computation times, which again limits operation under tight latency constraints.

Several works aimed at alleviating the harmful effect of stragglers without sacrificing their contributions.
Client selection protocols were proposed to account for expected latency~\cite{nishio2019client}, share calculation time between users and server~\cite{zhang2023timelyfl}, and assign lesser weight to older updates~\cite{hu2023scheduling}, while potentially reducing the effective data coverage per round and risking systemic bias if the same set of low-latency clients is repeatedly selected \cite{li2020federated}.
Other methods allow variable workload or model size on different devices by, e.g., training smaller local models on weaker devices \cite{diao2020heterofl}.
Recently, more refined strategies geared towards tight latency FL have leveraged the structure of deep networks by allowing partial model updates.
Notably, straggler-aware layer-wise federated learning (SALF)~\cite{lang2024stragglers} introduced layer-wise aggregation, where even slow clients contribute gradients for the layers they complete within the deadline.
This layer-wise approach is well-suited for deep neural networks where backpropagation is inherently sequential. While this improves utilization and convergence, existing methods like SALF typically assumed fixed deadlines and batch sizes, missing the potential benefits of dynamically allocating both per-round time and per-client workload based on system heterogeneity.

While prior work addresses straggler mitigation either at the round level (deadlines, client selection) or the local optimization level (layer-wise partial updates), there is no unified approach that optimizes both the \emph{deadline} and the \emph{local optimization} to best use the total available time.
Yet, these decisions are inherently coupled: longer deadlines enable more computation from stragglers, while local optimization procedure affects how much local data each client processes and how long each round takes.
Furthermore, existing methods such as SALF default to applying a full FedAvg update when no updates are available for a layer, a practice that can misalign the model under strict time constraints.

We propose \emph{Adaptive Deadline and Batch Layered Federated Learning} (ADEL-FL), a novel framework that jointly optimizes per-round deadlines in synchronous FL and the mini-batch sizes used for local learning, combined with layer-wise aggregation.
In contrast to existing methods, which either fix the deadline per round \cite{chen2016revisiting} or treat it as a heuristic hyperparameter \cite{bonawitz2019towards}, ADEL-FL actively schedules a different deadline for \emph{each round}, as part of a global optimization objective.
Specifically, ADEL-FL formulates the deadline assignment and mini-batch sizing as a constrained optimization problem aimed at minimizing the expected distance to the optimal model under total training time and communication round constraints.

\subsection*{Summary of Contributions}

\begin{itemize}[leftmargin=5pt]
    \item {\bf Optimized Straggler-Adaptive Synchronous FL Algorithm}: We introduce ADEL-FL, the first FL method to jointly optimize round deadlines and batch sizes with layer-wise aggregation, 
    balancing between reducing stochastic gradient variance and mitigating deadline-induced truncation variance leading to improved convergence under time constraints.
    Unlike prior methods, our approach accounts for iteration-varying deadlines and modified batch sizes, and it preserves layer parameters when no updates are received.
    \item {\bf Theoretical Analysis}: By modeling per-layer computation time using a realistic exponential distribution, which enables analytical estimates of expected updates under deadline constraints and guiding optimal allocation of time and workload, we provide convergence guarantees for ADEL-FL.
    We rigorously show that its partial updates are unbiased with bounded variance, and analyze its effect on convergence of the learning procedure.
    \item {\bf Extensive Experimentation}: We show experimentally that ADEL-FL achieves superior convergence and accuracy compared to existing straggler mitigation approaches under heterogeneous settings.
    For example, we show an improvement of over 19\% in validation accuracy for MNIST \cite{deng2012mnist} and over 13\% in CIFAR-10 \cite{krizhevsky2009learning} over existing methods.
\end{itemize}

The rest of the paper is structured as follows: Section \ref{sec:system_model} presents the system model and the problem formulation;
Section \ref{sec:method} details ADEL-FL; Section \ref{sec:experiments} provides experimental results; and Section \ref{sec:conclusion} concludes the paper.

\section{Preliminaries and Problem Formulation}\label{sec:system_model}
In this section, we lay the foundation for deriving ADEL-FL.
We begin with modeling synchronous FL in Section~\ref{subsec:FL}, after which we introduce the system heterogeneity model and discuss its impact on local computation and communication latency in Section~\ref{subsec:Heterogeneity}.
We then formulate the tackled problem in Section~\ref{subsec:problem_formulation}.

\subsection{Federated Learning} \label{subsec:FL}
In FL \cite{mcmahan2017communication}, a central server trains a model with $n$  parameters denoted $\vw \in \R^n$, utilizing data stored at a group of $U$ users indexed by $u \in \{1, ..., U\}$.
Unlike traditional centralized learning, their corresponding datasets---denoted $\cD_1, \cD_2, ..., \cD_U$---cannot be transferred to the server due to privacy or communication restrictions \cite{gafni2021federated}.
Letting $F_u(\vw)$ denote the empirical risk function of the $u$th user, FL aims to find the  parameters $\vw_{\rm opt}$ which minimize the averaged empirical risk across all users, i.e.,

\begin{equation} \label{eq:opt_params}
\vw_{\rm opt} = \arg\min_{\vw} \left\{ F(\vw) \triangleq \frac{1}{U} \sum_{u=1}^{U} F_u(\vw) \right\},
\end{equation}
where we further assumed, for simplicity, that the local datasets are all of the same cardinality.

Broadly speaking, FL follows an iterative procedure operated in rounds.
In the $t$th round, the server transmits the current global model $\vw_t$ to the participating users, who each updates the model using its local dataset and computational resources.
These updates are then broadcasted back to the server, which aggregates them to form the new global model;
and this process continues until desired convergence is reached. 
In particular, conventional FL involves training at the devices via local SGD~\cite{stich2018local} and server's aggregation using FedAvg~\cite{mcmahan2017communication}.
The former takes the form
\begin{equation}\label{eq:local_sgd}
    \vw^u_{t+1} \xleftarrow{}  \vw_t -\eta_t \nabla F_u\left(\vw_t;i_{t}^{u}\right),
\end{equation}
where $i_{t}^{u}$ is the batch index chosen uniformly from all batches of cardinality $S_{t}^{u}$ in $\cD_u$, and $\eta_t$ is the learning rate.
For FedAvg, the aggregation rule is given by
\begin{equation} \label{eq:FedAvg_update_rule}
\vw_{t+1} \triangleq \frac{1}{U} \sum_{u=1}^{U} \vw^u_{t+1}= \vw_{t} - \eta_{t} \frac{1}{U} \sum_{u=1}^{U} \nabla F_{u}(\vw_{t}; i_{t}^{u}).
\end{equation}

Once \eqref{eq:FedAvg_update_rule} is employed for {\em synchronous} FL~\cite{mcmahan2017communication}, the server has to `wait' for the updates to arrive from all participating devices.
This, in turn, can induce notable latencies and delays once the edge clients are highly computationally heterogeneous, as discussed next.

\subsection{Computation and Communication Heterogeneity} \label{subsec:Heterogeneity}
In FL, participating edge devices often exhibit significant variability in their computational and communication capabilities, resulting in different processing times for computing the local model updates (gradients) in~\eqref{eq:local_sgd} and uploading them to the server \cite{yang2020delay, asad2023limitations}.
Formally, considering the $u$th user at global round $t$, let $T_{t}^{{\rm b},u}$ be the computational time required for backpropagation and locally calculating its corresponding gradient $\nabla F_u\left(\vw_t;i_{t}^{u}\right)$.
The individual computational speed $T_{t}^{{\rm b},u}$ can dramatically alter with respect to both $u$ and $t$ due to hardware limitations or additional computational tasks being executed concurrently.
In addition, after completing the local computation, each device must transmit its updated model to the server in communication time denoted by $T_{t}^{{\rm c},u}$ at round $t$ for user $u$.
The total iteration time $T_{t}^{u}$ of user $u$ is given by: $T_{t}^{u} = T_{t}^{{\rm b},u} + T_{t}^{{\rm c},u}$. Consequently, the overall computational latency per round $t$ in synchronous FL is determined by the {\em slowest} device, namely, $\max_{u} T_{t}^{u}$.
This constraint can make synchronous FL abortive for time-sensitive applications, where rapid learning is essential.
This can be tackled, e.g., by imposing a global, per-round, fixed deadline $T_{t}^{\rm d}$ of maximum allowable computation time \cite{chen2016revisiting}.
Accordingly, nodes not meeting the deadline termed {\em stragglers} are dropped in aggregation such that \eqref{eq:FedAvg_update_rule} transforms into
\begin{equation}\label{eq:drop_stragglers_update_rule}
\begin{aligned}
\vw^{\rm drop}_{t+1} & = \vw^{\rm drop}_t - \eta_{t} \frac{1}{|\cU_{t}|} \sum_{u \in \cU_{t}} \nabla F_{u}(\vw^{\rm drop}_t; i_{t}^{u}), \\ \text{where} & \quad
\cU_t= \left\{ u \mid T_{t}^{u} \leq T_{t}^{\rm d} \right\}.
\end{aligned}
\end{equation}

SALF \cite{lang2024stragglers} improves upon \eqref{eq:drop_stragglers_update_rule} by, instead of dropping the stragglers, utilizes their partial gradients that are calculated within the deadline limitation, forming the global model in a layer-wise fashion.
In particular, for an $L$-layer neural network $\vw_t = \left[ \vw_t^1,\dots, \vw_t^l, \dots, \vw_t^L \right]^T$, and user updates $\vw_{u,t}^{l}$ \eqref{eq:drop_stragglers_update_rule} changes into updating each layer $l$ separately via
\begin{equation} \label{eq:SALF_stragglers_update_rule}
    \tilde{\vw}^{l}_{t+1} = 
    \begin{cases}
        \tilde{\vw}^{l}_{t}, & \text{if } |\cU_{t}^{l}| = 0, \quad \\
        \frac{1}{1 - p_{t}^{l}} 
        \left( \sum\limits_{u \in \cU_{t}^{l}} \frac{1}{|\cU_{t}^{l}|} \vw^{l}_{u,t} - p_{t}^{l} \tilde{\vw}^{l}_{t} \right) 
        & \text{otherwise},
    \end{cases}
\end{equation}
where $p_{t}^{l} \triangleq P\left( |\cU_{t}^{l}| = 0 \right)$ is a bias correction constant.

\subsection{Problem Formulation}\label{subsec:problem_formulation}
For low-latency and dynamic heterogeneous FL setting, the timing limitation is often static, rather than varying per-round \cite{li2023fedtcr}.
Additionally, the total number of global rounds is also commonly restricted, to limit the communication load \cite{park2021few}.
Thus, we aim to enhance straggler-aware FL by meeting the following requirements:
\begin{enumerate}[leftmargin=15pt,label={\em R\arabic*}]
    \item \label{itm:total_rounds} {\em \underline{Total rounds}}: The FL procedure cannot surpass a predefined number $R$ of global rounds.
    \item \label{itm:total_training_time} {\em \underline{Total training time}}: The total time required for learning the model parameters is at most $T_{\max}$.
    That is, if the execution time of the $t$th FL round is $T_{t}^{\rm d}$, then $\sum_{t=1}^{R} T_{t}^{\rm d} \leq T_{\max}$.
    \item \label{itm:device_hetrogenity} {\em \underline{Devices heterogeneity}}: Users differ in their computational capabilities, thus the latency of computing the gradients for a data batch can greatly vary between users.
\end{enumerate}
\section{ADEL-FL} \label{sec:method}
\begin{figure*}[t]
    \centering
    \includegraphics[width=0.9\linewidth]{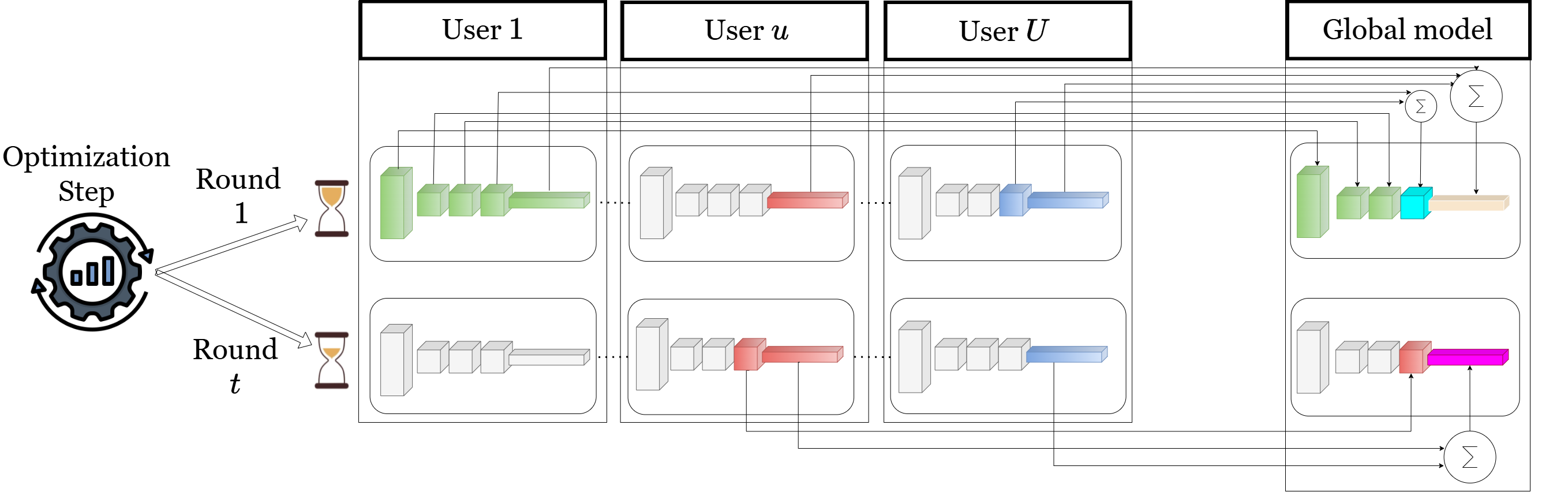}
    \caption{\small
     ADEL-FL illustration. Initially, the server executes an optimization step to determine the adaptive per-round deadlines $\{T_{t}^{\rm{d}}\}_{t=1}^{R}$, represented by the varying amount of sand in the hourglass. Then, for every round, each user performs depth-limited backpropagation, with the extent of computation governed by the deadline. The server aggregates layer-wise gradients across users using only the layers received before the deadline. The color of each layer reflects the set of users that contributed to it, visualized by blending the corresponding user colors. For example, if User $1$ is green and User $U$ is blue, a cyan-colored layer indicates that both contributed to that layer in the current round.
     }
    \label{fig:adel-system}
    \vspace{-0.4cm}
\end{figure*}

Evidently, Requirements \ref{itm:total_rounds}-\ref{itm:device_hetrogenity} can be satisfied by setting $T_{t}^{\rm d}=T_{\max}/R$, fixing the batch sizes, and employing layer-wise aggregation in~\eqref{eq:SALF_stragglers_update_rule}.
Nevertheless, the need to meet \ref{itm:total_rounds}-\ref{itm:device_hetrogenity} while achieving a high-performance FL model learned in dynamically changing environments, motivates an optimized choice of $T_{t}^{\rm d}$ to best leverage the available local resources per round $t$.
Moreover, as the local computation time of a user can be altered via its local training hyperparameters, and particularly its batch size, in this work, we seek a joint optimization of the per-round deadline and per-user batch size.
The proposed ADEL-FL scheme dynamically combines layer-wise aggregation as in \eqref{eq:SALF_stragglers_update_rule} with varying deadline and local optimization, as illustrated in Figure~\ref{fig:adel-system}.
To derive how ADEL-FL sets its deadlines and mini-batch sizes, in Section~\ref{sec:optimization_problem} we first formulate a constrained optimization problem using the expected $\ell_2$ distance.
Next, in Section \ref{sec:cost_function} we derive the bound on the $\ell_2$ distance and in Section \ref{sec:algorithm} we present the overall ADEL-FL algorithm.

\subsection{Optimization Problem}\label{sec:optimization_problem}
The formulation in Section~\ref{subsec:problem_formulation} motivates formulating a constrained optimization problem holding \ref{itm:total_rounds}-\ref{itm:device_hetrogenity}.
To cope with \ref{itm:device_hetrogenity}, we allow the batch sizes to vary between users, drawing inspiration from previous works \cite{ma2023adapbatch,liu2023adacoopt} that advocated the usage of different batch sizes to deal with heterogeneity.
Accordingly, we commence by determining a cost function that captures the model performance as a function of the deadlines $\{T_{t}^{\rm d}\}_{t=1}^{R}$ and the local batch sizes $\{S_{t}^{u}\}_{u=1,t=1}^{U, R}$.
To maintain tractability, we control all user batch sizes $\{S_{t}^{u}\}_{u=1,t=1}^{U, R}$ via a single global scaling parameter $m$. This parameter relates the local workload of each device to the round deadline, and will be formally defined in Section \ref{sec:cost_function}.

Following the common practice in the FL literature \cite{li2019convergence,wu2021fast,fraboni2023general}, assess the learned FL model via its expected $\ell_2$ distance to the optimal FL model as defined in Eq.~\eqref{eq:opt_params}.
Accordingly, by letting $\tilde{\vw}_{t}\left(\{T_{\tau}^{d}\}, m \right)$ denote the global model obtained at round $t$ via layer-wise aggregation \eqref{eq:SALF_stragglers_update_rule} with deadlines $\{T_{\tau}^{d}\}_{\tau \leq t}$ and mini-batch sizes $\{S_{t}^{u}\}_{u=1,t=1}^{U, R}$, we design ADEL-FL based on the following optimization framework:

\begin{problem} \label{problem:optimization}
For a total time budget $T_{\max}$ and a fixed number of rounds $R$, minimize the expected squared distance between $\tilde{\vw}_{R+1}$ and the global optimum $\vw_{opt}$:
\begin{equation*}
\begin{aligned}
& \min_{\{T_{t}^{\rm d}\}_{t=1}^{R}, m}  \mathbb{E}\left[\left\|\tilde{\vw}_{R+1} \left( \{T_{t}^{\rm d}\}_{t=1}^{R}, m \right)-\vw_{\rm opt}\right\|^2\right] \\
& \textrm{s.t.} \sum_{t = 1}^{R} T_{t}^{\rm d} \leq T_{\max}.
\end{aligned}
\end{equation*}
\end{problem}

\subsection{Deriving the Cost Function} \label{sec:cost_function}
 We now turn our attention to explicitly derive the cost function of Problem \ref{problem:optimization} for the layer-wise federated operation.
The modified operation of ADEL-FL means that existing FL analyses for scenarios with partial device participation (such as the one found in \cite{li2019convergence}) do not directly hold here.
Thus, to derive a concrete policy from Problem~\ref{problem:optimization}, we next lay out the assumptions that underpin our analysis, along with the statistical characteristics of the stragglers.
Following this, we present the derivation of the convergence bound. All proofs are delegated to the appendix.

\subsubsection{Assumptions}
We perform the derivation of the cost function under the following three assumptions, commonly adopted in FL studies~\cite{li2019convergence,stich2018local,lang2024stragglers}:

\begin{enumerate}[leftmargin=15pt,label={\em A\arabic*}]
  \item \label{itm:convex_smoothness} {\em \underline{Convexity and Smoothness}}: The local objectives $\left\{F_u(\cdot)\right\}_{u=1}^U$ are $\rho_c$-strongly convex and $\rho_s$-smooth.
  That is, for all $\vw_1, \vw_2 \in \R^n$, it holds that
\begin{equation*}
\begin{gathered}
\left(\vw_1-\vw_2\right)^T \nabla F_u\left(\vw_2\right)+\frac{1}{2} \rho_c\left\|\vw_1-\vw_2\right\|^2 \\
\leq F_u\left(\vw_1\right)-F_u\left(\vw_2\right) \leq \\
\left(\vw_1-\vw_2\right)^T \nabla F_u\left(\vw_2\right)+\frac{1}{2} \rho_s\left\|\vw_1-\vw_2\right\|^2.
\end{gathered}    
\end{equation*}  

\item \label{itm:variance_bound} {\em \underline{Variance Bound}}: For each user $u$ and round index $t$, the variance of the batch gradient $\nabla F_{u}\left(\vw ; i_{t}^{u}\right)$ is bounded by $\frac{\sigma_u^2}{S_{t}^{u}}$ for all $\vw \in \R^n$, i.e., 
\begin{equation*}
\mathbb{E}\left[\left\|\nabla F_u\left(\vw ; i_t^u\right)-\nabla F_u(\vw)\right\|^2\right] \leq \frac{\sigma_u^2}{S_{t}^{u}}.
\end{equation*}

\item \label{itm:gradient_bound} {\em \underline{Gradient Bound}}: For each user $u$ and round index $t$, the expected squared $\ell_2$ norm of the stochastic gradients $\nabla F_u\left(\vw ; i_t^u\right)$ is uniformly bounded by some $G^2$ for all $\vw \in \R^n$, i.e.,
\begin{equation*}
\mathbb{E}\left[\left\|\nabla F_u\left(\vw ; i_t^u\right)\right\|^2\right] \leq G^2.
\end{equation*}
\end{enumerate}

Under Assumption \ref{itm:variance_bound}, the variance of the batch gradient is inversely proportional to the batch size, referring to sampling with replacement of the samples.
This relation is crucial in FL scenarios, as it directly links the user-level batch size to stability and convergence guarantees. Such variance-batch proportionality is supported in prior works (e.g., \cite{stich2018local,li2019convergence}).
Beyond these fundamental analytical assumptions, we also specify the following modeling regarding batch sizes and computational latency, which are grouped separately to distinguish them from the core function properties:

\begin{enumerate}[leftmargin=15pt,label={\em B\arabic*}]

\item \label{itm:computational_model} {\em \underline{Computational Model}}: Let $T_{t,l}^{{\rm b},u}$ be the computation time of user $u$ in round $t$ for the backpropagation of layer $l$.
The random variables $\{T_{t,l}^{{\rm b},u}\}$ are i.i.d (in $u$, $t$, and $l$) and obey an exponential distribution:
\begin{equation*}
T_{t,l}^{{\rm b},u} \sim {\rm Exp}\left(\frac{S_{t}^{u}}{P_{u}}\right), 
\end{equation*}
so that $\mathbb{E}[T_{t,l}^{{\rm b},u}] = \frac{S_t^u}{P_u}$.

\item \label{itm:communicational model} {\em \underline{Communicational Model}}: We model the per-round communication time $T_{t}^{{\rm c},u}$ as a user-dependent, deterministic value that is constant across rounds:
\begin{equation*}
T_{t}^{{\rm c},u} = B_{u}   
\end{equation*}

\item \label{itm:batch_size} {\em \underline{Batch Size}}: The batch size of user $u$ in round $t$ is dictated by a global scaling parameter $m$, such that:
\begin{equation*}
S_{t}^{u} =  \Bigg\lfloor m P_{u} \left( \frac{T_{t}^{\rm d}-B_{u}}{T_{t}^{\rm d}} \right) \Bigg\rfloor.
\end{equation*}
\end{enumerate}
Model Formulation \ref{itm:computational_model} represents the computation time of a layer as an exponential random variable.
The exponential model is memoryless and well-suited for modeling compute time variability in deadline-constrained execution. This assumption aligns with prior work in straggler modeling and distributed training analysis~\cite{dutta2018slow,shi2020device,lee2017speeding}, where exponential or shifted-exponential distributions are used to enable tractable performance bounds and scheduling policies.
Model Formulation \ref{itm:communicational model} models the communication time for each client as a deterministic constant.
This is justified by the fact that the transmission time is dominated by stable bandwidth and latency characteristics of the client’s connection. While network fluctuations exist in practice, their variance is typically negligible compared to the overall training timescale \cite{jia2024efficient}.
Model Formulation \ref{itm:batch_size} ensures that faster devices do not dominate training by contributing excessively more updates within the same deadline, which could bias the model toward their data and hinder convergence to the global optimum $\vw_{\rm opt}$ in~\eqref{eq:opt_params} \cite{li2019convergence, lang2024stragglers, esfahanizadeh2022stream, zhao2018federated}.

In addition, we define the heterogeneity gap, quantifying the differences in data distribution  between users \cite{li2019convergence}:
\begin{equation} \label{eq:het_gap}
\Gamma \triangleq F\left(\vw_{\mathrm{opt}}\right)-\frac{1}{U} \sum_{u=1}^U \min _{\vw} F_u(\vw),
\end{equation}
where $\vw_{\rm opt}$ is defined in \eqref{eq:opt_params}.
A list of notation and symbols used throughout the analysis, including those defined in the preceding sections, is provided in Table \ref{tab:core_notation}.

\begin{table}[h]
\centering
\caption{Core Notation for ADEL-FL}
\label{tab:core_notation}
\begin{tabular}{|l|l|p{1cm}|}
\hline
\textbf{Symbol} & \textbf{Description} & \textbf{Defined in} \\
\hline
\multicolumn{3}{|c|}{\textbf{FedAvg}} \\
\hline
$U, u$ & index of the user &  \\
$R, t$ & index of the global round & \ref{itm:total_rounds}  \\
$L, l$ & index of the DNN layer &   \\
$T_{\max}$ & total training time budget & \ref{itm:total_training_time} \\
$T_{t}^{{\rm b},u}$ & backpropagation time for user $u$ for round $t$ &  \\
$T_{t}^{{\rm c},u}$ & communication time for user $u$ for round $t$ &  \\
$T_{t}^{u}$ & total iteration time for user $u$ for round $t$ &  \\
$S_{t}^{u}$ & batch size of the $u$th client at time $t$ & \\
$i_{t}^{u}$ & data sample index of the $u$th client at time $t$ &  \\
$\eta_t$ & learning rate at global round $t$ &  \\
$\nabla F_u$ & stochastic gradient of the $u$th client &  \\
$\vw_{\rm opt}$ & global minimizer of the cost function & Eq.~\eqref{eq:opt_params} \\
${\vw}_{t}$ & FedAvg global model for round $t$ & Eq.~\eqref{eq:FedAvg_update_rule}  \\
\hline
\multicolumn{3}{|c|}{\textbf{ADEL-FL}} \\
\hline
$T_{t}^{\rm d}$ & adaptive per-round deadline for round $t$ &  \\
$m$ & global batch-scaling parameter & \ref{itm:batch_size}  \\
$\cU_t^l$ & set of users contributing an update for layer $l$ & Eq.~\eqref{eq:drop_stragglers_update_rule} \\ 
$p_{t}^{l}$ & probability that $|\cU_t^l|=0$ & Eq.~\eqref{eq:SALF_stragglers_update_rule} \\
$\tilde{\vw}_{t}^{l}$ & ADEL-FL global layer $l$ update for round $t$ & Eq.~\eqref{eq:SALF_stragglers_update_rule} \\ 
$\vw^{l}_{u,t}$ & ADEL-FL user $u$ layer $l$ update for round $t$ &  \\ 
\hline
\multicolumn{3}{|c|}{\textbf{Analysis Parameters}} \\
\hline
$\rho_s, \rho_c$ & smoothness and convexity constants & \ref{itm:convex_smoothness} \\
$\sigma_{u}^{2}, G$ & variance and gradient bounds & \ref{itm:variance_bound}, \ref{itm:gradient_bound} \\
$P_{u}$ & computational capability of user $u$ & \ref{itm:computational_model} \\
$B_{u}$ & communication time for user $u$ & \ref{itm:communicational model} \\
$\Gamma$ & heterogeneity gap & Eq.~\eqref{eq:het_gap}  \\
\hline
\end{tabular}
\end{table}

\subsubsection{Convergence Bound}

Under the above assumptions and model formulations, we can now derive the bias correction factor $p_{t}^{l}$ defined in  \eqref{eq:SALF_stragglers_update_rule}:

\begin{lemma} \label{lemma:p_{t}^{l}}
When Assumptions \ref{itm:convex_smoothness}-\ref{itm:gradient_bound} and Model Formulations \ref{itm:computational_model}-\ref{itm:batch_size} hold, the value of $p_{t}^{l} \triangleq P(|\cU_{t}^{l}| = 0)$ is bounded by:
\begin{equation}
    p_{t}^{l} \leq  Q\left( L+1-l,  \frac{T_{t}^{\rm d}}{m} \right)^{U},
\end{equation}
where $ Q(s, x) \triangleq \frac{1}{\Gamma(s)} \int_x^{\infty} t^{s-1} e^{-t} dt$ is the regularized upper incomplete gamma function \cite{dlmf}.
\end{lemma}

As expected, $p_{t}^{l}$ is monotonically decreasing with the layer index $l$, as the recursive structure of the backpropagation algorithm implies that first layers are less likely to be reached under tight deadline.
Using Lemma \ref{lemma:p_{t}^{l}}, we state two lemmas for unbiasedness and bounded variance of each round of our proposed framework:

\begin{lemma} \label{lemma:unbiasedness}
(Unbiasedness).
Under Assumptions \ref{itm:convex_smoothness}-\ref{itm:gradient_bound} and Model Formulations \ref{itm:computational_model}-\ref{itm:batch_size}, for every FL round of index $t$ and a given set of training data samples $\{i_{t}^{u}\}$, the global model aggregated via ADEL-FL using \eqref{eq:SALF_stragglers_update_rule} is an unbiased estimator of the one obtained via vanilla FedAvg \cite{li2019convergence}, namely,
\begin{equation}
\mathbb{E}\left[\tilde{\vw}_{t+1}\right]=\vw_{t+1},
\end{equation}
where $\vw_{t+1}$ is given by \eqref{eq:FedAvg_update_rule}.
\end{lemma}

\begin{lemma} \label{lemma:bounded_var} (Bounded variance) Consider ADEL-FL with Assumptions \ref{itm:convex_smoothness}-\ref{itm:gradient_bound}, Model Formulations \ref{itm:computational_model}-\ref{itm:batch_size} and a given set of training data samples $\{i_{t}^{u}\}$. Given the learning rate $\eta_{t}$ set to be non-increasing and satisfying $\eta_{t} \leq 2\eta_{t+1}$, $T_{t+1}^{\rm d} \leq T_{t}^{\rm d}$ and $p_{t}^{1} < 0.2$, then, for every FL round $t$, the expected difference between $\tilde{\vw}_{t+1}$ and $\vw_{t+1}$, which is the variance of $\tilde{\vw}_{t+1}$, is bounded by:
\begin{equation}\label{maxQ}
\begin{gathered}
    \mathbb{E}\left[\left\|\tilde{\vw}_{t+1}-\vw_{t+1}\right\|^2\right] \leq \eta_t^2 G^2 \frac{4 U}{(U-1)} \times \\ \sum\limits_{l=1}^{L} \frac{1+Q\left( L+1-l,  \frac{T_{t}^{\rm d}}{m} \right)^{U}}{1-5Q\left( L+1-l,  \frac{T_{t}^{\rm d}}{m} \right)^{U}}.
\end{gathered}
\end{equation}
\end{lemma}

Lemmas \ref{lemma:unbiasedness} and \ref{lemma:bounded_var} are critical for establishing the convergence bound.
Lemma \ref{lemma:unbiasedness} ensures that ADEL-FL update remains an unbiased estimator of the FedAvg update, preserving convergence direction.
Lemma \ref{lemma:bounded_var} guarantees that the variance introduced by partial updates remains controlled. These properties mirror the core assumptions required in classical FL convergence analyses, such as those in \cite{li2019convergence}, and are necessary to derive a meaningful upper bound on the expected distance to $\vw_{\rm opt}$.
We are now ready to state the main theoretical contribution of this work: a novel convergence bound for the ADEL-FL framework that, for the first time, characterizes the fundamental interplay between adaptive per-round deadlines and batch-size scaling.
Unlike prior frameworks that treat round timing and local workload as static or decoupled parameters, our result provides a unified analytical objective that captures the trade-off between stochastic gradient noise and deadline-induced truncation variance.

\begin{theorem} \label{theorem:convergence}
Consider ADEL-FL with Assumptions \ref{itm:convex_smoothness}-\ref{itm:gradient_bound}, Model Formulations \ref{itm:computational_model}-\ref{itm:batch_size} and a given set of training data samples $\{i_{t}^{u}\}$. Given the learning rate $\eta_{t}$ is set to be non-increasing and holds $ \eta_{t}\leq 2\eta_{t+1}$, $\eta_{1} \leq \frac{1}{4\rho_{s}}$, $T_{t+1}^{\rm d} \leq T_{t}^{\rm d}$ and $p_{t}^{1} < 0.2$. Then, we have: 
\begin{equation}
\begin{gathered}
\mathbb{E}\left[\left\|\tilde{\vw}_{R+1}-\vw_{\rm opt}\right\|^2\right] \leq \prod_{t=1}^{R} \left( 1 - \eta_{t}\rho_{c} \right) \Delta_{1} 
\\ + \sum_{t=1}^{R} \eta_{t}^{2} \left(B_t + C_{t} \right) 
\prod_{\tau = t+1}^{R} \left( 1- \eta_{\tau}\rho_{c} \right),
\end{gathered}
\end{equation}
where $\Delta_{1}\triangleq \mathbb{E}\left[\left\|\tilde{\vw}_{1}-\vw_{opt}\right\|^2\right]$, and $B_t$ and $C_{t}$ are given by:
\begin{equation}
\begin{aligned}
B_t & \triangleq \frac{1}{U^{2}} \sum_{u=1}^{U} \frac{\sigma_{u}^{2}}{ m P_{u} \left( \frac{T_{t}^{\rm d} - B_{u}}{T_{t}^{\rm d}} \right) - 1 } +6 \rho_s \Gamma \\ C_{t} & \triangleq G^{2} \frac{4U}{U-1} \sum\limits_{l=1}^{L} \frac{1+Q\left( L+1-l,  \frac{T_{t}^{\rm d}}{m} \right)^{U}}{1-5Q\left( L+1-l,  \frac{T_{t}^{\rm d}}{m} \right)^{U}}.
\end{aligned}
\end{equation}
\end{theorem}

The convergence bound is governed by the constants $B_t$ and $C_t$, which represent the stochastic gradient variance and the deadline-induced truncation variance, respectively. Specifically, $B_t$ captures errors originating from data heterogeneity and finite batch sampling, while $C_t$ accounts for the information loss when users fail to complete deep network layers within the allotted time. These terms exhibit a fundamental tension mediated by the batch-scaling factor $m$: a larger $m$ reduces $B_t$ by improving update quality, but simultaneously exacerbates $C_t$ because the increased computational load leads to higher layer-wise truncation under a fixed deadline. This trade-off is analyzed in detail in Section~\ref{sec:discussion} and serves as a principled, tractable objective for the adaptive scheduling framework formulated in Problem~\ref{problem:optimization}.

\subsection{ADEL-FL Algorithm} \label{sec:algorithm}

Having derived the cost function in Theorem~\ref{theorem:convergence}, we are now ready to present the complete ADEL-FL algorithm, summarized in Algorithm~\ref{algorithm:adel_fl}.
In the beginning of the learning process (Line \ref{line:problem}), the server solves the global optimization problem derived from Theorem~\ref{theorem:convergence} to jointly determine the round-wise deadlines $\{T_{t}^{\rm d}\}_{t=1}^R$ and the global batch-scaling parameter $m$:

\begin{problem} \label{problem:optimization_full}
Given the convergence bound in Theorem~\ref{theorem:convergence}, minimize the upper bound on the expected squared $\ell_2$ distance between $\tilde{\vw}_{R+1}$ and the global optimum $\vw_{opt}$ under a total time budget $T_{\max}$:
\begin{equation*}
\begin{aligned}
 \min_{\{T_{t}\}_{t=1}^{R}, m} &  \prod_{t=1}^{R} \left( 1 - \eta_{t}\rho_{c} \right) \Delta_{1} \\ 
 + & \sum_{t=1}^{R} \eta_{t}^{2} \left(B_t + C_{t} \right) \prod_{\tau = t+1}^{R} \left( 1- \eta_{\tau}\rho_{c} \right)
\\ \textrm{s.t.} & \sum_{t = 1}^{R} T_{t}^{\rm d} \leq T_{\max}. \\ 
& T_{t+1}^{\rm d} \leq T_{t}^{\rm d}, \quad \forall t \in \{1, \dots, R-1\}, \\
 & p_{t}^{1} < 0.2 \quad \forall t \in \{1, \dots, R\}.
\end{aligned}
\end{equation*}
\end{problem}

The server solves Problem~\ref{problem:optimization_full} using a trust-region method~\cite{yuan2000review}, an iterative strategy well-suited for non convex constrained optimization.
Trust-region approaches approximate the objective by a local quadratic model and restrict updates to a neighborhood where the model is deemed reliable.
The region's size is adaptively adjusted based on how well the approximation predicts true improvement.
Subsequently, the server computes each user's batch size $ S_{t}^{u} $ via Assumption \ref{itm:batch_size} proportional to their computational capability (Line \ref{line:su}).
Once the initial configuration is complete, the server transmits the initial global model $\vw_1$, the round-specific deadlines $T_{t}^{\rm d}$, and the local batch sizes $S_{t}^{u}$ to each participating client.

In each learning round $t$, users perform local training within their time limit, computing gradients up to a layer $d_t^u$ based on how far they progress before the effective deadline $T_{t}^{\rm d} - B_{u}$ (Line \ref{line:gradient}), and then transmit those partial updates to the server (Line \ref{line:covney}).
Following the expiration of the deadline, the server initiates a layer-wise aggregation process to form the new global model update.
For each layer $l \in \{1, \dots, L\}$, the server identifies the subset of participating users $\mathcal{U}_t^l$ who successfully transmitted their gradients for that specific layer before the cutoff (Line \ref{line:recover}).
The aggregation is performed using the bias-corrected rule in Eq.~\eqref{eq:SALF_stragglers_update_rule}, which explicitly compensates for the partial participation induced by the deadline to ensure the update remains an unbiased estimator of the full gradient (Line \ref{line:compute}). This procedure repeats for $R$ rounds, and the final model is returned in Line \ref{line:output}.

\begin{algorithm}
\small
\caption{ADEL-FL}\label{algorithm:adel_fl}
\begin{algorithmic}[1]
\State \textbf{Input:} Number of rounds $R$, training deadline $T_{\max}$, learning rates $\left\{ \eta_{t} \right\}_{t=1}^{R}$, computational capabilities $\{P_{u}\}_{u=1}^{U}$, communication times $\{B_{u}\}_{u=1}^{U}$ and initial model $\vw_1$ \label{line:input}
\Indent
\State Solve Problem \ref{problem:optimization_full} to find $\{T_{t}^{\rm d}\}_{t=1}^{R}, m$ \label{line:problem} 
\State Set batch sizes $S_{t}^{u}$ for each $u\in\{1,\ldots,U\}$ and $t\in\{1,\ldots,R\}$  via Model Formulation \ref{itm:batch_size} \label{line:su}
\For{$t = 1, \dots, R$}
\State \underline{Clients} - do in parallel for each $u$, until deadline $T_{t}^{\rm d} - B_{u}$: \label{line:client}
\Indent
\State Compute $\nabla F_u\left(\vw_t, i_t^u\right)$ up to layer $d_t^u$ \label{line:gradient}
\State Convey partial gradients to server \label{line:covney}
\EndIndent
\State \underline{Server} - do after deadline time $T_{t}^{\rm d}$: \label{line:server}
\Indent
\For{$l = 1, \dots, L$}
\State Recover the participating user set $\cU_t^l$ \label{line:recover}
\State Compute $\tilde{\vw}_{t+1}^l$ via Eq.~ \eqref{eq:SALF_stragglers_update_rule} \label{line:compute}
\EndFor
\EndIndent
\EndFor
\EndIndent
\State \textbf{Output:} Trained model $\vw_{R+1}$ \label{line:output}
\end{algorithmic}
\end{algorithm}

The overall strategy of ADEL-FL refines the convergence guarantees derived earlier by optimally tuning the time-budgeted training configuration.
By solving Problem~\ref{problem:optimization_full}, we adaptively allocate deadlines and batch sizes to each round, enabling balanced participation and efficient learning.
The resulting algorithm demonstrates improved performance under strict time constraints, as empirically demonstrated in Section \ref{sec:experiments}.

\subsection{Discussion} \label{sec:discussion}
The ADEL-FL framework aims to minimize convergence error by adaptively managing the interplay between local workload and global timing constraints.
This is achieved by optimizing the trade-off between stochastic gradient noise and partial participation variance, as formalized in the cost function of Problem \ref{problem:optimization_full}. The cost function reflects a fundamental trade-off between the terms $B_t$ and $C_t$, which capture distinct sources of error.
The term $B_t$ corresponds to the variance induced by stochastic gradients, and is a monotonically decreasing function of the global batch scaling factor $ m $.
Thus, increasing $ m $ (i.e., using larger batches) reduces $B_t$, reducing the user's SGD variance.
However, larger batches require more computation per user, potentially reducing the number of clients able to contribute full updates within a fixed deadline.
This effect is captured by $ C_t $, which quantifies the variance due to partial participation and depends on the regularized incomplete gamma function.
As $ m $ increases, $C_t$ worsens due to fewer layers being computed per user under time constraints as in Requirement \ref{itm:total_training_time}.
In addition, the per-round deadline $ T_{t}^{\rm d} $ must be judiciously allocated across rounds to align with the learning rate schedule and batch size configuration, ensuring that each round maximally leverages the available computational budget while preserving convergence efficiency.
A natural extension of ADEL-FL is to also optimize the number of rounds $R$, in addition to $T_{t}^{\rm d}$ and $m$.
In practical deployments, both the number of rounds and the overall latency are often constrained.
Optimizing over $R$ introduces an additional degree of flexibility, allowing the system to determine the optimal number of global rounds given the available time budget.
This joint optimization problem (over the number of rounds $R$, per-round deadlines $\{T_{t}^{\rm d}\}$, and batch scaling $m$) could be formulated as a mixed-integer constrained program or tackled with adaptive scheduling heuristics, replacing Problem \ref{problem:optimization_full} in Step \ref{line:problem} of the algorithm.
Incorporating this optimization into ADEL-FL would further enhance its applicability to time-critical FL scenarios.
Furthermore, while this work assumes a deterministic communication model for tractability, incorporating the extension to stochastic or time-varying communication models remains an important direction for future research.

\begin{figure*}
    \centering
    \includegraphics[width=0.85\linewidth]{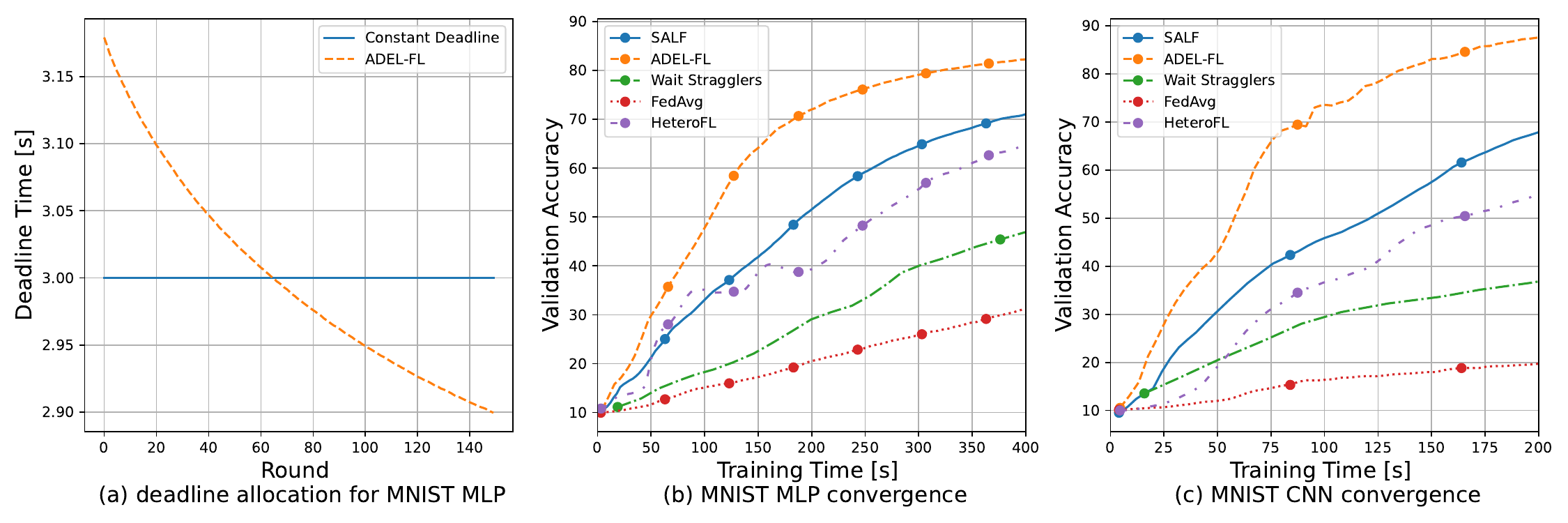}
    \caption{\small Deadline allocation and convergence curves for  MNIST  using an inverse decaying learning rate. $(a)$ Adaptive deadline per round for ADEL-FL (MLP).$(b)$~Convergence, MLP. $(c)$ Adaptive deadline per round for ADEL-FL (CNN).}
    \label{figure:mnist}
    \vspace{-0.4cm}
\end{figure*}

\section{Experimental Study} \label{sec:experiments}
We empirically evaluate ADEL-FL for training image classification models using MNIST~\cite{deng2012mnist} and CIFAR-10~\cite{krizhevsky2009learning}.
We consider a series of experiments of increasing complexity. In each case, we compare ADEL-FL against four baseline schemes: 
$(i)$ {\em Wait Stragglers:} the standard federated averaging with full client participation (synchronous FedAvg with the server waiting for all devices each round)~\cite{mcmahan2017communication};
$(ii)$ {\em Drop-Stragglers:} synchronous FedAvg with a fixed per-round deadline that drops any late client updates~\cite{chen2016revisiting};
$(iii)$ {\em SALF:} the straggler-aware FL method suggested in ~\cite{lang2024stragglers}, which aggregates partial gradients from slow devices but uses a fixed deadline and batch size;
and $(iv)$ {\em HeteroFL:} which trains local submodels with reduced width according to each client's capability, allowing efficient aggregation into a full global model~\cite{diao2020heterofl}.
All methods are evaluated under a fixed total training time budget $T_{\max}$ and a fixed number of communication rounds $R$ (per the constraints in Requirements~\ref{itm:total_rounds}–\ref{itm:total_training_time}).
Experiments are conducted on a Linux server with NVIDIA A30 GPU.
To emulate heterogeneous device speeds in a controlled manner, we sample each client’s computation time per layer from an exponential distribution (as in Model Formulation~\ref{itm:computational_model}).
We begin with image classification tasks on MNIST and CIFAR-10 using an inverse-decay learning rate schedule $\eta_{t} = \eta_{0}/(1+t)$ with initial $\eta_{0}\in\{0.05,\,0.1,\,0.5,\,1.0\}$ to assess baseline convergence behavior.
Next, we analyze stability and optimization dynamics under constant learning rate and adding $\ell_2$ regularization, deviating from the assumptions of Theorem~\ref{theorem:convergence}.
Finally, we present a comparative table summarizing performance across varying total training times, highlighting ADEL-FL's efficiency under strict time constraints.

\subsection{MNIST} We first evaluate on the MNIST dataset~\cite{deng2012mnist} (handwritten $28\times 28$ grayscale digits).
We set the total training time $T_{\max}$ and number of rounds $R$ such that, under a standard local mini-batch size, the average fraction of each model updated per round (i.e. the average backpropagation depth) is $50\%$ of the network layers.
We consider two model architectures of different capacity: $(1)$ MLP: a simple fully-connected network with two hidden layers (32 and 16 neurons) and a softmax output;
and $(2)$ CNN: a small CNN with two $5\times5$ convolutional layers (each followed by max-pooling and ReLU) and two dense layers, ending in a softmax output.

Figure~\ref{figure:mnist}(a) shows that the traditional constant deadline allocation versus the proposed deadline allocation which decreases over time, similarly to the learning rate, optimizing early contributions from slower devices.
Figure~\ref{figure:mnist}(b) and (c) demonstrates that ADEL-FL achieves faster and higher convergence than baselines while dynamically adapting deadlines to gradually accelerate training.
ADEL-FL substantially outperforms \emph{Drop-Stragglers}, \emph{Wait Stragglers} and \emph{HeteroFL}, which suffer from slow or stalled convergence.
\emph{SALF} mitigates stragglers but remains less efficient than ADEL-FL, particularly in early training, where larger deadlines allow more layers to be updated.
For example, ADEL-FL achieves over 20\% accuracy improvement on MLP at time $180[s]$ compared to \emph{SALF}.
\vspace{-0.2cm}
\subsection{CIFAR-10} 
\begin{figure}[t]
    \centering
    \includegraphics[width=\linewidth]{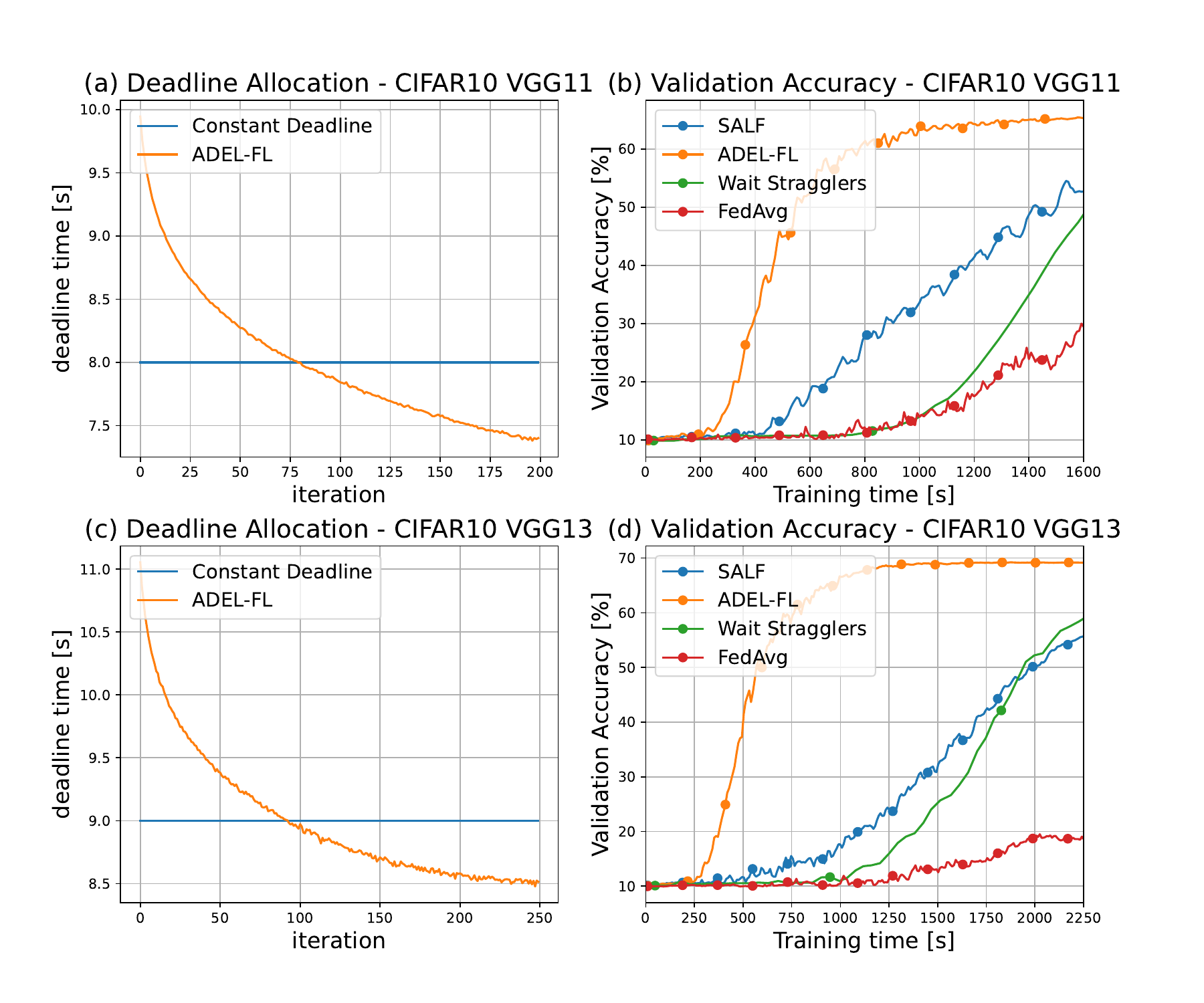}
    \vspace{-0.8cm}
    \caption{\small Deadline allocation and convergence curves for CIFAR-10 using an inverse decaying learning rate. $(a)$ Adaptive deadline per round for ADEL-FL (VGG11).$(b)$~Convergence, VGG11. $(c)$ Adaptive deadline per round for ADEL-FL (VGG11).$(d)$~Convergence, VGG11.}
    \label{figure:cifar}
    \vspace{-0.4cm}
\end{figure}

Next, we evaluate on CIFAR-10, simulating an FL system with $U=30$ users.
The data are partitioned across the users using a Dirichlet distribution with concentration parameter $\alpha = 0.5$, following the standard non-IID construction of \cite{hsu2019measuring}.
This procedure samples client-specific label proportions from a Dirichlet prior, producing heterogeneous yet controllable data heterogeneity across users.
Consequently, each client receives a distinct subset of the 60{,}000 training images that reflects a moderately non-IID allocation of the global dataset. In these experiments, we focus on deeper neural network models.
We train two architectures of the VGG family~\cite{simonyan2015very}: VGG11 and VGG13, which have 8 and 10 convolutional layers (respectively) followed by 3 fully-connected layers.
These networks are substantially deeper than the MNIST models, providing a more challenging and heterogeneous workload.
We set the global time budget $T_{\max}$ and rounds $R$ such that the average local computation per round reaches about $85\%$ of each model’s full depth for a baseline batch size, i.e., on average clients nearly complete a full pass through the network each round under the time limit.
The resulting allocations and validation accuracies are reported in  Figure~\ref{figure:cifar}.
We observe that ADEL-FL consistently outperforms \emph{SALF}, \emph{Drop-Stragglers}, and \emph{Wait Stragglers}, achieving faster convergence while maintaining a higher final accuracy (66\% for ADEL-FL vs 53\% for SALF).
As in MNIST, the deadline allocation decreases over time alongside the decaying learning rate, efficiently balancing training speed and convergence.
\vspace{-0.2cm}
\subsection{Robustness to Learning Assumptions}
\begin{figure}[t]
    \centering
    \includegraphics[width=\linewidth]{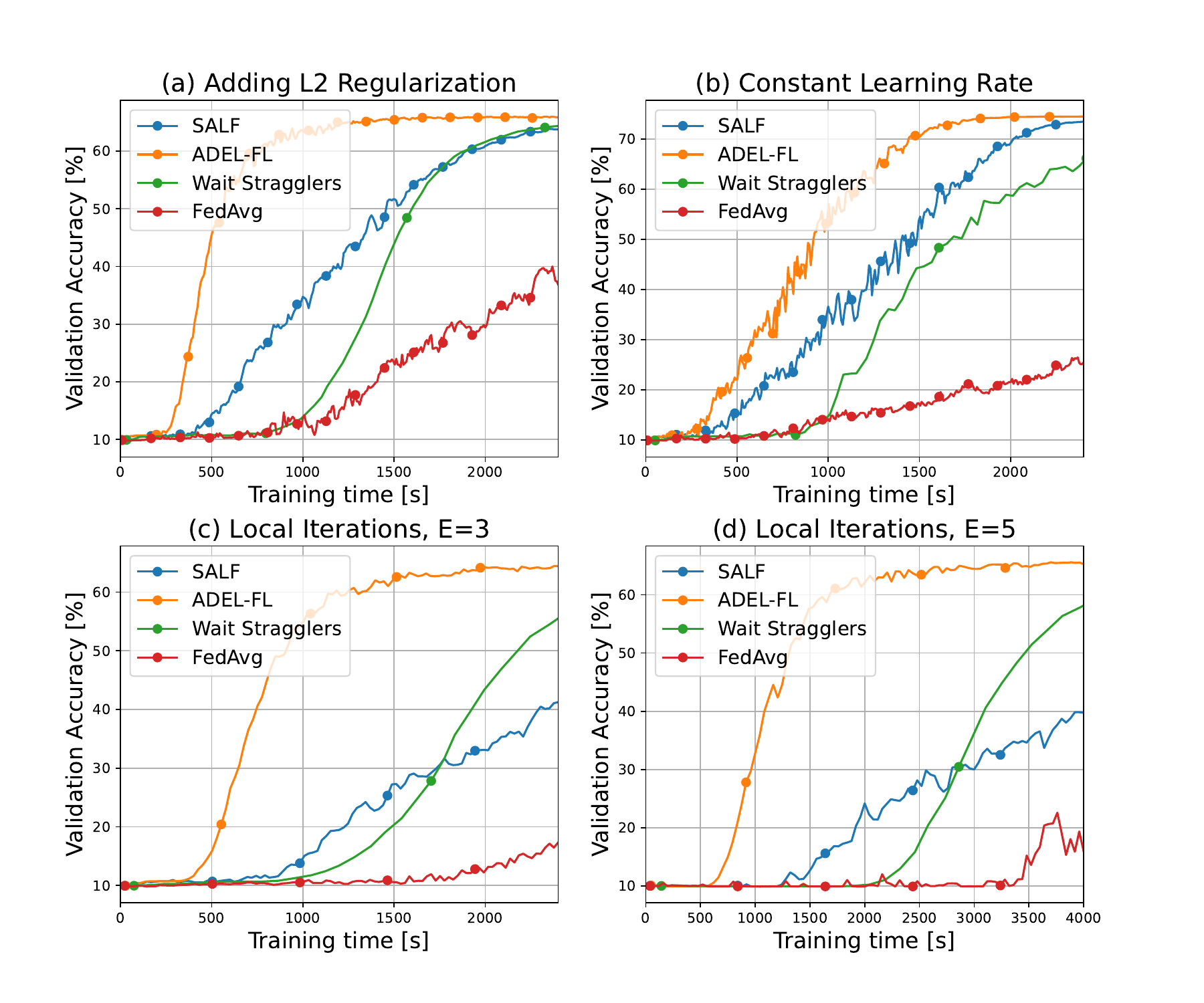}
    \vspace{-0.8cm}
    \caption{\small Robustness results, CIFAR-10 VGG11. $(a)$ Convergence with $\ell_2$ regularization. $(b)$ Convergence with constant learning rate. $(c)$ Convergence with $E=3$ local iterations. $(d)$ Convergence with $E=5$ local iterations.}
    \label{figure:aux}
    \vspace{-0.4cm}
\end{figure}

To validate ADEL-FL under conditions where optimization assumptions are violated, we perform four studies on CIFAR-10 VGG11, reported in Figure~\ref{figure:aux}.
First, we introduce $\ell_2$ regularization on local objectives. Using $\ell_2$ regularization has been shown to improve generalization and to stabilize training in modern deep networks, including when adaptive optimizers are used.
Empirical studies demonstrate that decoupled weight decay improves final accuracy and yields more reliable behavior across datasets and architectures, which makes it a practical tool to mitigate overfitting on clients with limited local data \cite{loshchilov2019decoupled, mcmahan2017communication, li2020federated}.
Second, we fix the learning rate $\eta_t=\eta_0$ instead of using a decaying profile.
As discussed in \cite{goodfellow2016deep}, maintaining a constant learning rate during early optimization helps SGD preserve gradient noise, enabling broader exploration of the loss landscape and reducing the tendency to converge prematurely to sharp or suboptimal minima. 

So far, we have focused our formulation and experiments on settings where the users compute a single stochastic gradient over the selected mini-batch, i.e., using the local SGD terminology as in~\cite{stich2018local},  
by letting $E$ be the number of local SGD iterations, we so far used $E=1$.
In the third and fourth experiments, we vary the number of local SGD iterations before each global update to $E=3$ and $E=5$ respectively.
Increasing the number of local iterations $E$ is a logical and widely used strategy in FL because it reduces communication frequency, lowers network load, and allows each global aggregation step to amortize more local computation.
Prior work has shown that performing multiple local updates per round significantly decreases communication requirements in practice \cite{mcmahan2017communication}, and that local SGD can maintain strong convergence behavior even under aggressive communication reduction \cite{stich2018local}.
At the same time, larger $E$ increases the risk of client drift in heterogeneous data regimes \cite{karimireddy2020scaffold}.
These modifications deviate from theoretical conditions assumed for convergence in Theorem~\ref{theorem:convergence}.
Despite the fact that the considered setting no longer aligns with the assumptions based on which our objective is derived, we observe in Figure~\ref{figure:aux} that ADEL-FL maintains advantages over \emph{SALF}, \emph{Wait-Stragglers} and \emph{Drop-Stragglers}.
Across all stress tests, the method exhibits stable descent behavior and improved final accuracy, suggesting that its synchronization and deadline-adaptive aggregation mechanism compensates for moderate deviations from canonical optimization assumptions.
The convergence remains stable and outperforms the other existing solutions compared, demonstrating that ADEL-FL's design is robust even when the actual optimization dynamics differs moderately from the assumed model.
This robustness further indicates that ADEL-FL can be deployed in realistic federated environments where heterogeneous client behaviors often violate idealized assumptions, without sacrificing stability or efficiency.

\subsection{Training Time} 
We conclude our experimental study by evaluating the performance achieved under predefined overall deadlines.
Table~\ref{tab:training-time-sweep} compares final accuracy across different total training time budgets for the VGG11 architecture on the CIFAR-10 dataset, using an IID data partition across clients.
Even under stringent time limits, ADEL-FL preserves superior convergence, highlighting its efficient use of limited computation.
As the total training time $T_{\max}$ increases, all methods generally benefit from improved model accuracy due to deeper local computations.
However, ADEL-FL consistently maintains a noticeable gap over \emph{SALF}, \emph{Wait-Stragglers} and \emph{Drop-Stragglers} across all budgets.
In the low-budget regime, ADEL-FL is particularly advantageous, achieving much better performance while effectively handling the computation resources.
In higher budget regimes, although the gap slightly narrows, ADEL-FL continues to converge faster and achieves superior final performance.

\begin{table}[ht]
\small
\centering
\caption{\small Final test accuracy (\%) of ADEL-FL and baseline methods on CIFAR-10 (VGG11) under varying total training time budgets $T_{\max}$.}
\label{tab:training-time-sweep}
\begin{tabular}{lcccc}
\toprule
\textbf{$T_{\max} [s]$} & \textbf{ADEL-FL} & \textbf{SALF} & \textbf{Drop-Stragglers} & \textbf{FedAvg} \\
\midrule
1200 & \textbf{61} & 36 & 14 & 30 \\
1600 & \textbf{66} & 50 & 16 & 46 \\
2000 & \textbf{68} & 61 & 17 & 57 \\
2400 & \textbf{69} & 66 & 20 & 63 \\
\bottomrule
\end{tabular}
\end{table}

\section{Conclusion} \label{sec:conclusion}

We introduced ADEL-FL, a novel FL framework that jointly optimizes per-round deadlines and client-specific batch sizes under a global training time constraint.
By integrating adaptive scheduling with layer-wise aggregation, ADEL-FL effectively mitigates stragglers while preserving statistical efficiency.
Our theoretical analysis establishes that ADEL-FL yields unbiased updates with bounded variance under exponential latency models, and our optimization formulation captures the convergence trade-offs induced by time allocation and batch scaling.
Empirical results on standard benchmarks demonstrate that ADEL-FL consistently outperforms existing methods in convergence rate and final accuracy under heterogeneous settings.

\appendix
\section{Technical Appendices and Supplementary Material}

\subsection{Appendix A - Proof of Lemma \ref{lemma:p_{t}^{l}}} \label{Subsec:AppendixPl}
\numberwithin{equation}{section}
\setcounter{equation}{0}

We initiate the proof by defining $d_{t}^{u}$ as the index of the final layer reached during the backpropagation process by user $u$ in round $t$.
Given a model with total depth $L$, this index is determined by the relationship:
\begin{equation}
    d_{t}^{u} = \max\{L+1 - z_{t}^{u}, 1\},
\end{equation}
where $z_{t}^{u}$ represents the number of layers that would be processed within the effective deadline $T_{t}^{\rm d} - B_{u}$ assuming an unconstrained (potentially infinite-depth) sequence of layers.
This formulation is used since $z_{t}^{u}$ is stochastic and unbounded, while the actual model depth is limited to $L$.
Under Model Formulation \ref{itm:computational_model}, backpropagation can be modeled as sequential layer gradient computation with exponential service time.
By the fundamental relationship between the exponential distribution and the Poisson process \cite{ross2014introduction}, the number of layers processed $z_t^u$ in the fixed interval $[0, T_{t}^{\rm d} - B_u]$ follows a Poisson distribution:
\begin{equation}
z_t^u \sim \text{Poiss}\left( \lambda_{t}^{u} \right), \quad \text{where } \lambda_{t}^{u} = \frac{P_u}{S_{t}^{u}} (T_{t}^{\rm d} - B_u).
\end{equation}
By substituting $S_{t}^{u} =  \Bigg\lfloor m P_{u} \left( \frac{T_{t}^{\rm d} - B_{u}}{T_{t}^{\rm d}} \right) \Bigg\rfloor$ from Model Formulation \ref{itm:batch_size} we get: 
\begin{equation*}
    \lambda_{t}^{u} = \frac{m P_{u} \left( \frac{T_{t}^{\rm d} - B_{u}}{T_{t}^{\rm d}} \right)}{\Bigg\lfloor m P_{u} \left( \frac{T_{t}^{\rm d} - B_{u}}{T_{t}^{\rm d}} \right) \Bigg\rfloor} \frac{T_{t}^{\rm d}}{m}
\end{equation*} 

Recall that $p_{t}^{l} \triangleq P(|\cU_{t}^{l}| = 0)$ denotes the probability that no user contributes an update for layer $l$ at round $t$.
We use the i.i.d (by $u$) property of Model Formulation \ref{itm:computational_model} to write:
\begin{equation*}
\begin{aligned}
    P(|\cU_{t}^{l}| = 0) & = \bigcap_{u=1}^{U} P(z_{t}^{u} \leq L-l) \\ & = \prod_{u=1}^{U} P(z_{t}^{u} \leq L-l) = \left( P(z_{t}^{1} \leq L-l) \right)^{U}
\end{aligned}
\end{equation*}

To simplify the bound, we introduce an auxiliary random variable $\tilde{z}_{t}$:
\begin{equation}
    \tilde{z}_{t} \sim \text{Poiss}\left(\tilde{\lambda}_t\right), \quad \text{where } \tilde{\lambda}_t = \frac{T_{t}^{\rm d}}{m},
\end{equation}
which provides an upper bound on the zero-contribution probability:
\begin{equation}
    P(|\cU_{t}^{l}| = 0) = \left( P(z_{t}^{1} \leq L-l) \right)^{U} \leq \left( P(\tilde{z}_{t} \leq L-l) \right)^{U}.
\end{equation}
This inequality holds because $\lambda_t^u \geq \tilde{\lambda}_t$, and since the Poisson cumulative distribution function is a monotonically decreasing function of its rate, the smaller rate $\tilde{\lambda}_t$ provides a conservative upper bound for the zero-contribution probability. The probability mass function (PMF) of $X \sim \text{Poiss}(\lambda)$ is given by:
\begin{equation*}
    P(X = k) = \frac{\lambda^k e^{-\lambda}}{k!}, \quad \text{for} \quad k = 0,1,2,\dots
\end{equation*}

We get:
\begin{equation} \label{equation:pl_sum}
    P(|\cU_{t}^{l}| = 0) \leq \left( P(\tilde{z}_{t} \leq L-l) \right)^{U} = \left( \sum_{k=0}^{L-l} \frac{\tilde{\lambda}_{t}^{k} e^{-\tilde{\lambda}_{t}}}{k!} \right)^{U}.
\end{equation}

It can be observed that the summation term in Eq.~\eqref{equation:pl_sum} coincides with the regularized upper incomplete gamma function evaluated at a specific argument.
The following auxiliary lemma formalizes this identity, with the proof deferred to Section E.

\textbf{Auxiliary Lemma.} For an integer $s \ge 1$,
\begin{equation}
Q(s,x) = \sum_{k=0}^{s-1} \frac{x^k e^{-x}}{k!},
\end{equation}
where $Q(s,x) \triangleq \frac{1}{\Gamma(s)} \int_{x}^{\infty} t^{s-1} e^{-t} dt$ is the regularized upper incomplete gamma function.

Using the auxiliary lemma with $s-1 = L-l$ and $x = \frac{T_{t}^{\rm{d}}}{m}$, Eq.~\eqref{equation:pl_sum} can be written as

\begin{equation*}
    p_{t}^{l} = P(|\cU_{t}^{l}| = 0) \leq Q\left( L+1-l,  \frac{T_{t}^{\rm d}}{m} \right)^{U},
\end{equation*}
which completes the proof of Lemma \ref{lemma:p_{t}^{l}}. \hfill $\square$

\subsection{Appendix B - Proof of Lemma \ref{lemma:unbiasedness}} \label{Subsec:AppendixMean}
We aim to demonstrate that $\tilde{\vw}_{t+1}$, as defined in Eq.~\eqref{eq:SALF_stragglers_update_rule}, serves as an unbiased estimator of $\vw_{t+1}$ from Eq.~\eqref{eq:FedAvg_update_rule}.
To achieve this, we analyze the expectation of the $l$-th subvectors of both expressions, leveraging the decomposition of the gradient into layers.
Given the batches $\{i_u^t\}$, the only randomness in $\vw_{t+1}^l$ originates from $\cU_{t}^{l}$.
Instead of directly computing the expectation over $\cU_{t}^{l}$, we use the law of total expectation:

\begin{equation*}
\begin{aligned}
    \mathbb{E}\Big[\tilde{\vw}_{t+1}^l\Big] & = \mathbb{E} \Big[ \mathbb{E} \Big[\tilde{\vw}_{t+1}^l \Big| |\cU_{t}^{l}| \Big] \Big] \\
    & = \sum_{K=0}^{U} P(|\cU_{t}^{l}| = K) \mathbb{E} \Big[\tilde{\vw}_{t+1}^l \Big| |\cU_{t}^{l}| = K\Big].
\end{aligned}
\end{equation*}

Substituting the probability of $|\cU_t^l| = 0$, denoted as $p_{t}^{l}$, we get:
\begin{equation*}
\begin{aligned}
    \mathbb{E}[\tilde{\vw}_{t+1}^l] & = p_{t}^{l} \tilde{\vw}_t^l + \sum_{K=1}^{U} P(|\cU_t^l| = K) \frac{1}{1 - p_{t}^{l}} \cdot \\ & \left( \frac{1}{K} \mathbb{E} \left[ \sum_{u \in \cU_t^l} \vw_{u,t}^l \middle|  |\cU_t^l| = K \right] - p_{t}^{l} \tilde{\vw}_t^l \right).
\end{aligned}
\end{equation*}

By Lemma 4 in \cite{li2019convergence}, when users are uniformly sampled without replacement, FedAvg provides an unbiased estimate of full device participation.
Hence, we can write:

\begin{equation} \label{AppendixMean:fedavg_mean}
    \frac{1}{K} \mathbb{E} \left[ \sum_{u \in \cU_t^l} \vw_{u,t}^l \middle|  |\cU_t^l| = K \right] = \frac{1}{K} K \sum_{u=1}^{U} \frac{1}{U} \vw_{u,t}^l = \vw_{t+1}^l.
\end{equation}

Substituting this into our expectation computation and simplifying using the summation of probabilities, we obtain:
\begin{equation*}
    \mathbb{E}[\tilde{\vw}_{t+1}^l] = p_{t}^{l} \tilde{\vw}_t^l + \frac{\vw_{t+1}^l - p_{t}^{l} \tilde{\vw}_t^l}{1-p_{t}^{l}}\sum_{K=1}^{U} P(|\cU_t^l| = K).
\end{equation*}

Using:
\begin{equation*}
    \sum_{K=1}^{U} P(|\cU_t^l| = K) = 1-P(|\cU_t^l| = 0) = 1-p_{t}^{l},
\end{equation*}
we get:
\begin{equation*}
    \mathbb{E}[\tilde{\vw}_{t+1}^l] = p_{t}^{l} \tilde{\vw}_t^l + \vw_{t+1}^l - p_{t}^{l} \tilde{\vw}_t^l = \vw_{t+1}^l.
\end{equation*}

Since this equality holds for every layer $l \in \{1, \dots, L\} $, and the global model $ \tilde{\vw}_{t+1} $ is formed by stacking the layer-wise vectors $ \tilde{\vw}^l_{t+1} $, it follows that the full model update is also unbiased:
\begin{equation*}
\mathbb{E}[ \tilde{\vw}_{t+1} ] = \vw_{t+1},
\end{equation*}
completing the proof. \hfill $\square$

\subsection{Appendix C - Proof of Lemma \ref{lemma:bounded_var}} \label{Subsec:AppendixVar}

For each layer $l$ and round $t$, we define the variance of the layer update as $v_{t}^{l} \triangleq \mathbb{E}\Big[\|\tilde{\vw}_{t}^{l} - \vw_{t}^{l}\|^2\Big]$.
Using the law of total expectation with respect to the random variable $|\cU_{t}^l|$:
\begin{equation}
\label{AppendixVar:total_var}
\begin{aligned}
    v_{t+1}^{l} & \triangleq \mathbb{E}\Big[\|\tilde{\vw}_{t+1}^{l} - \vw_{t+1}^{l}\|^2\Big] \\
    & = P\Big[|\cU_{t}^{l}|=0\Big] \mathbb{E}\Big[\|\tilde{\vw}_{t+1}^{l} - \vw_{t+1}^{l}\|^2\ \Big| |{\cU}_{t}^{l}|=0\Big] \\
    & + \sum_{K=1}^{U} P\Big[|\cU_{t}^{l}|=K\Big] \mathbb{E}\Big[\|\tilde{\vw}_{t+1}^{l} - \vw_{t+1}^{l}\|^2 \,\Big|\, |\cU_{t}^{l}|=K\Big].
\end{aligned}
\end{equation}

For the first summand, it holds that:
\begin{equation} \label{AppendixVar:first_sum}
\begin{aligned}
    & \mathbb{E}\Big[\|\tilde{\vw}_{t+1}^{l} - \vw_{t+1}^{l}\|^2\ \Big| |{\cU}_{t}^{l}|=0\Big]
    =  \mathbb{E}\Big[\|\tilde{\vw}_{t}^{l} - \vw_{t+1}^{l}\|^2\Big] \\
    =\;& \mathbb{E}\Big[\|\tilde{\vw}_{t}^{l} - \vw_{t}^{l} + \vw_{t}^{l} - \vw_{t+1}^{l}\|^2\Big] \\
    =\;& \mathbb{E}\Big[\|\tilde{\vw}_{t}^{l} - \vw_{t}^{l}\|^2\Big]
    + \mathbb{E}\Big[\|\vw_{t+1}^{l} - \vw_{t}^{l}\|^2\Big] \\
    &\quad + 2\mathbb{E}\Big[\big(\tilde{\vw}_{t}^{l} - \vw_{t}^{l}\big)^T\big(\vw_{t+1}^{l} - \vw_{t}^{l}\big)\Big] \\
    \stackrel{(a)}{=}\;& v_t^l + \mathbb{E}\Big[\|\vw_{t+1}^{l} - \vw_{t}^{l}\|^2\Big] \\
    \stackrel{(b)}{\leq}\;& v_t^l + \sum_{u=1}^{U}\frac{1}{U}\mathbb{E}\Big[\|\vw_{u,t}^{l} - \vw_{t}^{l}\|^2\Big] \\
    =\;& v_t^l + \sum_{u=1}^{U}\frac{1}{U}\mathbb{E}\Big[\|\eta_t \nabla F_u(\vw_t;i_t^u)\|^2\Big]
    \stackrel{(c)}{\leq}\; v_t^l + \eta_t^2 G^2.
\end{aligned}
\end{equation}
where (a) follows by conditioning on the current batches $\{i_t^u\}_{u=1}^{U}$. Given $\{i_t^u\}_{u=1}^{U}$, the quantity $\vw_{t+1}^l-\vw_t^l$ is deterministic, and since $\tilde{\vw}_t^l$ is determined by the aggregation randomness up to round $t-1$ and is independent of the current batches, Lemma~\ref{lemma:unbiasedness} implies $\mathbb{E}\big[\tilde{\vw}_t^l-\vw_t^l \,\big|\, \{i_t^u\}_{u=1}^{U}\big] = \mathbf{0}$, which yields $\mathbb{E}\Big[\big(\tilde{\vw}_{t}^{l} - \vw_{t}^{l}\big)^T\big(\vw_{t+1}^{l} - \vw_{t}^{l}\big)\Big] = 0$.
Step (b) follows from the convexity of $\|\cdot\|^2$, and (c) follows from Assumption~\ref{itm:gradient_bound}.

For the second summand:
\begin{align}
& \mathbb{E}\Big[\|\tilde{\vw}_{t+1}^{l} - \vw_{t+1}^{l}\|^2\ \Big| |{\cU}_{t}^{l}|=K\Big] \notag \\
= & \mathbb{E}\Bigg[\Bigg|\Bigg|\frac{1}{1 - p_{t}^{l}} \left( \sum\limits_{u \in \cU_{t}^{l}} \frac{1}{|\cU_{t}^{l}|} \vw_{u,t}^l - p_{t}^{l} \tilde{\vw}_t^l \right) -\vw_{t+1}^{l}\Bigg|\Bigg|^2 \Bigg| |\cU_{t}^{l}|=K \Bigg] \notag \\    
= & \frac{1}{(1-p_{t}^{l})^2} \times \Bigg( \mathbb{E}\Bigg[\Bigg|\Bigg| \sum\limits_{u \in \cU_{t}^{l}} \frac{1}{|\cU_{t}^{l}|} \vw_{u,t}^l - \vw_{t+1}^{l} \notag \\
& -p_{t}^{l}(\tilde{\vw}_{t}^{l} - \vw_{t+1}^{l}) \Bigg|\Bigg|^2 \Bigg| |\cU_{t}^{l}|=K \Bigg] \Bigg) \notag \\ 
= &\frac{1}{(1-p_{t}^{l})^2} \times \Bigg( \mathbb{E}\Bigg[\Bigg|\Bigg| \sum\limits_{u \in \cU_{t}^{l}} \frac{1}{|\cU_{t}^{l}|} \vw_{u,t}^l - \vw_{t+1}^{l} \Bigg|\Bigg|^2 \Bigg| |\cU_{t}^{l}|=K \Bigg] \label{AppendixVar:fedavg_var} \\ 
& + (p_{t}^{l})^{2} \mathbb{E}\Big[\|\tilde{\vw}_{t}^{l} - \vw_{t+1}^{l}\|^2 \Big] \label{AppendixVar:null_update_var} - 2p_{t}^{l}(\tilde{\vw}_{t}^{l} - \vw_{t+1}^{l})^{T} \\
& \cdot \mathbb{E}\Bigg[ \left( \sum\limits_{u \in \cU_{t}^{l}} \frac{1}{|\cU_{t}^{l}|} \vw_{u,t}^l - \vw_{t+1}^{l} \right) \Bigg| |\cU_{t}^{l}|=K \Bigg] \Bigg).\label{AppendixVar:zeros_var}
\end{align}

The value in Eq.~\eqref{AppendixVar:fedavg_var} can be bounded using the Lemma 5 of \cite{li2019convergence}. That is, when $K$ out of $U$ users are uniformly sampled without replacement, the variance of the scheme compared to full users participation is bounded by:
\begin{equation*}
\begin{aligned}
& \mathbb{E}\Bigg[\Bigg|\Bigg| \sum\limits_{u \in \cU_{t}^{l}} \frac{1}{|\cU_{t}^{l}|} \vw_{u,t}^l - \vw_{t+1}^{l} \Bigg|\Bigg|^2 \Bigg| |\cU_{t}^{l}|=K \Bigg] \\
&\leq \frac{U}{K(U-1)} \left( 1 - \frac{K}{U} \right) 4\eta_{t}^{2}G^{2}.
\end{aligned}
\end{equation*}

The value in Eq.~\eqref{AppendixVar:null_update_var} is bounded using the results from Eq.~\eqref{AppendixVar:first_sum}, thus:
\begin{equation*}
    (p_{t}^{l})^{2} \mathbb{E}\Big[\|\tilde{\vw}_{t}^{l} - \vw_{t+1}^{l}\|^2\Big] \leq (p_{t}^{l})^{2} \left( v_{t}^{l} + \eta_{t}^{2}G^{2} \right).
\end{equation*}

Finally, using the result for the mean of $K$ user's update uniformly sampled without replacement in Eq.~\eqref{AppendixMean:fedavg_mean}, the value in Eq.~\eqref{AppendixVar:zeros_var} is equal to 0.
Adding the three expressions together, we get:
\begin{equation}
\label{AppendixVar:second_sum}
\begin{aligned}
& \mathbb{E}\Big[\|\tilde{\vw}_{t+1}^{l} - \vw_{t+1}^{l}\|^2\ \Big| |{\cU}_{t}^{l}|=K\Big] \leq \frac{1}{(1-p_{t}^{l})^2} \times \\ 
& \left( (p_{t}^{l})^{2} (\eta_{t}^{2}G^2 + v_{t}^{l}) + \frac{U}{K(U-1)}\left(1 - \frac{K}{U} \right) 4\eta_{t}^{2}G^{2} \right).
\end{aligned}
\end{equation}

Plugging Eq.~\eqref{AppendixVar:first_sum}~and Eq.~\eqref{AppendixVar:second_sum}~into Eq.~\eqref{AppendixVar:total_var}, we obtain:
\begin{equation*}
\begin{aligned}
v_{t+1}^{l} & = \mathbb{E}\Big[\|\tilde{\vw}_{t+1}^{l} - \vw_{t+1}^{l}\|^2\Big] \leq P\Big[|\cU_{t}^{l}|=0\Big] \left( v_{t}^{l} + \eta_{t}^{2}G^{2} \right) \\
& + \sum_{K=1}^{U} P\Big[|\cU_{t}^{l}|=K\Big] \frac{1}{(1-p_{t}^{l})^2} \\
& \left( (p_{t}^{l})^{2} (\eta_{t}^{2}G^2 + v_{t}^{l}) + \frac{U}{K(U-1)}\left(1 - \frac{K}{U} \right) 4\eta_{t}^{2}G^{2} \right) \\
& \leq p_{t}^{l} \left( v_{t}^{l} + \eta_{t}^{2}G^{2} \right) + \frac{1}{(1-p_{t}^{l})^2} \\ 
& \left( (p_{t}^{l})^{2} (\eta_{t}^{2}G^2 + v_{t}^{l}) + \frac{4\eta_{t}^{2}G^{2}U}{U-1}\right) \sum_{K=1}^{U} P\Big[|\cU_{t}^{l}|=K\Big] \\
& = p_{t}^{l} \left( v_{t}^{l} + \eta_{t}^{2}G^{2} \right) \\ 
& + \frac{1}{1-p_{t}^{l}} \left( (p_{t}^{l})^{2} (\eta_{t}^{2}G^2 + v_{t}^{l}) + \frac{4\eta_{t}^{2}G^{2}U}{U-1}\right) \\
& = \frac{p_{t}^{l}}{1-p_{t}^{l}} v_{t}^{l} + \eta_{t}^{2}G^{2} \left( \frac{\frac{4U}{U-1} + p_{t}^{l}}{1-p_{t}^{l}} \right).
\end{aligned}
\end{equation*}

This yields a recursive upper bound on the per-layer update variance $v^l_{t+1}$ as a function of the previous round's variance $v^l_t$:
\begin{equation}\label{eq:v_1}
    v_{t+1}^{l} \leq D_{t}^{l}\cdot v_{t}^{l} + E_{t}^{l} , \quad v_{1}^{l} = 0,
\end{equation}
where 
\begin{equation}\label{eq:E_and_D}
D_{t}^{l} \triangleq \frac{p_{t}^{l}}{1-p_{t}^{l}} \quad \text{and} \quad E_{t}^{l} \triangleq \eta_{t}^{2}G^{2} \left( \frac{\frac{4U}{U-1} + p_{t}^{l}}{1-p_{t}^{l}} \right).
\end{equation}

Since $T_{t}^{\rm d} \geq T_{t+1 }^{\rm d}$, we get $p_{t}^{l} \leq p_{t+1}^{l}$ and $D_{t}^{l} \leq D_{t+1}^{l}$. In addition, since the learning rate satisfies $\eta_{t} \leq 2\eta_{t+1}$, it follows that $\eta_{t}^{2} \leq 4\eta_{t+1}^{2}$ and therefore $E_{t}^{l} \leq 4E_{t+1}^{l}$. Unfolding the recursion yields:
\begin{equation*}
\begin{aligned}
v_{t+1}^{l} & \leq D_{t}^{l}\cdot v_{t}^{l} + E_{t}^{l} \leq D_{t}^{l} D_{t-1}^{l} v_{t-1}^{l} + D_{t}^{l}E_{t-1}^{l} + E_{t}^{l} \\ & \leq \sum_{k=1}^{t} E_{k}^{l} \prod_{j=k+1}^{t} D_{j}^{l} + \underbrace{v_{1}^{l}\prod_{k=1}^{t}D_{k}^{l}}_{0} \leq \sum_{k=1}^{t} E_{k}^{l} (D_{t}^{l})^{t-k}
 \\ & \leq \sum_{k=1}^{t} 4^{t-k} E_{t}^{l} (D_{t}^{l})^{t-k} = E_{t}^{l} \sum_{k=1}^{t}  (4D_{t}^{l})^{t-k} \\ & = \frac{1-(4D_{t}^{l})^{t}}{1-4D_{t}^{l}} E_{t}^{l},    
\end{aligned}
\end{equation*}
where the last equality holds since, by the conditions of Lemma~\ref{lemma:bounded_var}, we have $p_{t}^{l} \leq p_{t}^{1} < 0.2$. Substituting this bound into the definition of $D_{t}^{l}$ in Eq.~\eqref{eq:E_and_D} implies $D_{t}^{l} < 0.25$, and therefore $4D_{t}^{l} < 1$. Substituting $D_{t}^{l}$ and $E_{t}^{l}$ given in Eq.~\eqref{eq:E_and_D}, we have:
\begin{equation*}
\begin{aligned}
v_{t+1}^{l} & = \mathbb{E}\Big[\|\tilde{\vw}_{t+1}^{l} - \vw_{t+1}^{l}\|^2\Big] \leq \frac{1-(4D_{t}^{l})^{t}}{1-4D_{t}^{l}} E_{t}^{l} \\
& \leq \frac{E_{t}^{l}}{1-4D_{t}^{l}} = \eta_{t}^{2}G^{2} \left( \frac{\frac{4U}{U-1} + p_{t}^{l}}{1-5p_{t}^{l}} \right).
\end{aligned}
\end{equation*}

Now, summing over the $L$ layers:
\begin{equation*}
\begin{aligned}
    & \mathbb{E}\Big[\|\tilde{\vw}_{t+1} - \vw_{t+1}\|^2\Big] = \sum_{l=1}^{L} \mathbb{E}\Big[\|\tilde{\vw}_{t+1}^{l} - \vw_{t+1}^{l}\|^2\Big]  \\
    & \leq \sum_{l=1}^{L} \eta_{t}^{2}G^{2} \left( \frac{\frac{4U}{U-1} + p_{t}^{l}}{1-5p_{t}^{l}} \right) \leq \eta_{t}^{2}G^{2}\frac{4U}{U-1} \sum_{l=1}^{L} \frac{1 + p_{t}^{l}}{1-5p_{t}^{l}}.
\end{aligned}
\end{equation*}

Substituting the bound on $p_{t}^{l}$, as given in Lemma \ref{lemma:p_{t}^{l}}, gives:
\begin{equation*}
\begin{aligned}
&\mathbb{E}\left[\left\|\tilde{\vw}_{t+1}-\vw_{t+1}\right\|^2\right] \\
&\leq \eta_t^2 G^2 \frac{4 U}{(U-1)} \sum\limits_{l=1}^{L} \frac{1+Q\left( L+1-l,  \frac{T_{t}^{\rm d}}{m} \right)^{U}}{1-5Q\left( L+1-l,  \frac{T_{t}^{\rm d}}{m} \right)^{U}},    
\end{aligned}
\end{equation*}
thus proving Lemma \ref{lemma:bounded_var}. \hfill$\square$

\subsection{Appendix D - Proof of Theorem \ref{theorem:convergence}}
\label{Subsec:AppendixTheorem}

We start by using the full participation FedAvg update $\vw_{t+1}$:
\begin{equation*}
\begin{aligned}
     & \mathbb{E}\Big[\|\tilde{\vw}_{t+1} - \vw_{opt} \|^2 \Big] \\
     = & \mathbb{E}\Big[\|\tilde{\vw}_{t+1} - \vw_{t+1} + \vw_{t+1} 
    - \vw_{opt} \|^2 \Big] \\
     = & \underbrace{\mathbb{E}\Big[\|\tilde{\vw}_{t+1} - \vw_{t+1} \|^2 \Big]}_{S_1} + \underbrace{\mathbb{E}\Big[\|\vw_{t+1} - \vw_{opt} \|^2 \Big]}_{S_2} \\
     + & \underbrace{2\langle \tilde{\vw}_{t+1} - \vw_{t+1} , \vw_{t+1} 
    - \vw_{opt} \rangle }_{S_3}.
\end{aligned}
\end{equation*}

We note that given the batches $\{i_{t}^{u}\}$, $S_3$ vanishes by the unbiasedness Lemma \ref{lemma:unbiasedness}. $S_1$ is bounded using Lemma \ref{lemma:bounded_var}:
\begin{equation*}
\begin{aligned}
& \mathbb{E}\left[\left\|\tilde{\vw}_{t+1}-\vw_{t+1}\right\|^2\right] \\
\leq & \eta_t^2 G^2 \frac{4 U}{(U-1)} \sum\limits_{l=1}^{L} \frac{1+Q\left( L+1-l,  \frac{T_{t}^{\rm d}}{m} \right)^{U}}{1-5Q\left( L+1-l,  \frac{T_{t}^{\rm d}}{m} \right)^{U}}.
\end{aligned}
\end{equation*}

The term $S_2$ is bounded using Lemma 1 from \cite{li2019convergence}, which states that under Assumption~\ref{itm:convex_smoothness} and a step size satisfying $\eta_t \leq \frac{1}{4\rho_{s}}$, the following inequality holds:
\begin{equation}\label{eq:aveE}
\begin{aligned}
\mathbb{E}\Big[\|\tilde{\vw}_{t+1}^{l} - \vw_{opt} \|^2 \Big] & = (1-\eta_{t}\rho_{c}) \mathbb{E}\Big[\|\tilde{\vw}_{t}^{l} - \vw_{opt} \|^2 \Big] \\
& + \underbrace{\eta_{t}^{2} \mathbb{E}\Big[\|g_{t}-\bar{g}_{t} \|^2 \Big]}_{\tilde{F}}
+ 6\eta_{t}^{2}\rho_{s}\Gamma,    
\end{aligned}
\end{equation}
where $g_{t}$ and $\bar{g}_{t}$ are defined by:
\begin{equation*}
g_{t} \triangleq \frac{1}{U}\sum_{u=1}^{U}\nabla F_u\left(\vw ; i_t^u \right) \quad \bar{g}_{t} \triangleq \frac{1}{U}\sum_{u=1}^{U}\nabla F_u\left(\vw\right).
\end{equation*}

The quantity $\tilde{F}$ in Eq.~\eqref{eq:aveE} corresponding to the variance of the stochastic gradient error, admits the following upper bound:
\begin{equation*}
\begin{aligned}
    \eta_{t}^{2} \mathbb{E}\Big[\|g_{t}-\bar{g}_{t} \|^2 \Big] & = \eta_{t}^{2} \mathbb{E} \Bigg|\Bigg| \sum_{u=1}^{U} \frac{1}{U} \left( F_u\left(\vw ; i_t^u \right) - \nabla F_u\left(\vw\right) \right) \Bigg|\Bigg|^{2} \\
    & = \eta_{t}^{2}\frac{1}{U^{2}} \sum_{u=1}^{U} \mathbb{E} \Big|\Big| \left( F_u\left(\vw ; i_t^u \right) - \nabla F_u\left(\vw\right) \right) \Big|\Big|^{2} \\
    & = \eta_{t}^{2}\frac{1}{U^{2}} \sum_{u=1}^{U} \frac{\sigma_{u}^{2}}{S_{t}^{u}} \\
    & = \eta_{t}^{2}\frac{1}{U^{2}} \sum_{u=1}^{U} \frac{\sigma_{u}^{2}}{\left\lfloor m P_{u} \left( \frac{T_{t}^{\rm d} - B_{u}}{T_{t}^{\rm d}} \right) \right\rfloor} \\ 
    & \leq \eta_{t}^{2}\frac{1}{U^{2}} \sum_{u=1}^{U} \frac{\sigma_{u}^{2}}{ m P_{u} \left( \frac{T_{t}^{\rm d} - B_{u}}{T_{t}^{\rm d}} \right) - 1 },
\end{aligned}
\end{equation*}
where the third equality stems from Assumption \ref{itm:variance_bound}, the forth equality stems from Model Formulation \ref{itm:batch_size} and the inequality from $\frac{1}{\lfloor x \rfloor } \leq \frac{1}{x-1}, \quad \forall x >1$.

Combining the bounds for $S_1$ and $S_2$, we obtain:
\begin{equation*}
\begin{aligned}
    \mathbb{E}\Big[\|\tilde{\vw}_{t+1} - \vw_{opt} \|^2 \Big] & \leq (1-\eta_{t}\rho_{c}) \mathbb{E}\Big[\|\tilde{\vw}_{t}^{l} - \vw_{opt} \|^2 \Big] \\
    & + \eta_{t}^{2} \left( B_t + C_{t} \right),    
\end{aligned}
\end{equation*}
where $B_t$ and $C_{t}$ are given by:
\begin{equation*}
\begin{aligned}
B_t & \triangleq \frac{1}{U^{2}} \sum_{u=1}^{U} \frac{\sigma_{u}^{2}}{ m P_{u} \left( \frac{T_{t}^{\rm d} - B_{u}}{T_{t}^{\rm d}} \right) - 1 } +6 \rho_s \Gamma \\
C_{t} & \triangleq G^{2} \frac{4U}{U-1} \sum\limits_{l=1}^{L} \frac{1+Q\left( L+1-l,  \frac{T_{t}^{\rm d}}{m} \right)^{U}}{1-5Q\left( L+1-l,  \frac{T_{t}^{\rm d}}{m} \right)^{U}}.
\end{aligned}
\end{equation*}

By repeatedly applying the recursive relation from the final model $\vw_{R+1}$, we obtain:
\begin{equation*}
\begin{aligned}
\mathbb{E}\left[\left\|\tilde{\vw}_{R+1}-\vw_{opt}\right\|^2\right] & \leq \prod_{t=1}^{R} \left( 1 - \eta_{t}\rho_{c} \right) \Delta_{1} \\
& + \sum_{t=1}^{R} \eta_{t}^{2} \left(B_t + C_{t} \right) 
\prod_{\tau = t+1}^{R} \left( 1- \eta_{\tau}\rho_{c} \right)
\end{aligned}
\end{equation*}

where $\Delta_{1} \triangleq \mathbb{E}\left[\left\|\tilde{\vw}_{1}-\vw_{opt}\right\|^2\right] $, thus proving Theorem \ref{theorem:convergence}. \hfill$\square$

\subsection{Appendix E - Proof of Auxiliary Lemma} We derive this expression using integration by parts. For an integer $s > 1$:

\begin{align*}
u &= \frac{t^{s-1}}{\Gamma(s)}, & dv &= e^{-t} \, dt, \\
du &= \frac{t^{s-2}}{\Gamma(s-1)} \, dt, & v &= -e^{-t}.
\end{align*}

Applying integration by parts, we have:
\begin{align*}
Q(s,x) &= \left[-\frac{t^{s-1}e^{-t}}{\Gamma(s)} \right]_{t=x}^\infty + \frac{1}{\Gamma(s-1)}\int_{x}^\infty t^{s-2} e^{-t} dt \\
&= \frac{x^{s-1} e^{-x}}{\Gamma(s)} + Q(s-1, x) \\
&= \frac{x^{s-1} e^{-x}}{(s-1)!} + Q(s-1, x).
\end{align*}

Unfolding the recursion with stop condition $Q(1,x) = e^{-x}$ yields:
\begin{equation} \label{equation:gamma_and_sum}
    Q(s,x) = \sum_{k=0}^{s-1} \frac{x^k e^{-x}}{k!}, 
\end{equation} 
concluding the proof. \hfill$\square$

\bibliographystyle{IEEEtran} 
\bibliography{IEEEfull,biblio}

\begin{thebibliography}{10}
\providecommand{\url}[1]{#1}
\csname url@samestyle\endcsname
\providecommand{\newblock}{\relax}
\providecommand{\bibinfo}[2]{#2}
\providecommand{\BIBentrySTDinterwordspacing}{\spaceskip=0pt\relax}
\providecommand{\BIBentryALTinterwordstretchfactor}{4}
\providecommand{\BIBentryALTinterwordspacing}{\spaceskip=\fontdimen2\font plus
\BIBentryALTinterwordstretchfactor\fontdimen3\font minus \fontdimen4\font\relax}
\providecommand{\BIBforeignlanguage}[2]{{%
\expandafter\ifx\csname l@#1\endcsname\relax
\typeout{** WARNING: IEEEtran.bst: No hyphenation pattern has been}%
\typeout{** loaded for the language `#1'. Using the pattern for}%
\typeout{** the default language instead.}%
\else
\language=\csname l@#1\endcsname
\fi
#2}}
\providecommand{\BIBdecl}{\relax}
\BIBdecl

\bibitem{mcmahan2017communication}
B.~McMahan, E.~Moore, D.~Ramage, S.~Hampson, and B.~A. y~Arcas, ``Communication-efficient learning of deep networks from decentralized data,'' in \emph{Artificial intelligence and statistics}.\hskip 1em plus 0.5em minus 0.4em\relax PMLR, 2017, pp. 1273--1282.

\bibitem{zhou2021survey}
\BIBentryALTinterwordspacing
J.~Zhou, S.~Zhang, Q.~Lu, W.~Dai, M.~Chen, X.~Liu, S.~Pirttikangas, Y.~Shi, W.~Zhang, and E.~Herrera-Viedma, ``A survey on federated learning and its applications for accelerating industrial internet of things,'' 2021. [Online]. Available: \url{https://arxiv.org/abs/2104.10501}
\BIBentrySTDinterwordspacing

\bibitem{abdulrahman2020survey}
S.~Abdulrahman, H.~Tout, H.~Ould-Slimane, A.~Mourad, C.~Talhi, and M.~Guizani, ``A survey on federated learning: The journey from centralized to distributed on-site learning and beyond,'' \emph{IEEE Internet of Things Journal}, vol.~8, no.~7, pp. 5476--5497, 2021.

\bibitem{gafni2021federated}
T.~Gafni, N.~Shlezinger, K.~Cohen, Y.~C. Eldar, and H.~V. Poor, ``Federated learning: A signal processing perspective,'' \emph{IEEE Signal Processing Magazine}, vol.~39, no.~3, pp. 14--41, 2022.

\bibitem{stich2018local}
S.~U. Stich, ``Local {SGD} converges fast and communicates little,'' in \emph{International Conference on Learning Representations}, 2019.

\bibitem{yang2020delay}
Z.~Yang, M.~Chen, W.~Saad, C.~S. Hong, M.~Shikh-Bahaei, H.~V. Poor, and S.~Cui, ``Delay minimization for federated learning over wireless communication networks,'' in \emph{International Conference on Machine Learning, Workshop on Federated Learning}, 2020.

\bibitem{asad2023limitations}
\BIBentryALTinterwordspacing
M.~Asad, S.~Shaukat, D.~Hu, Z.~Wang, E.~Javanmardi, J.~Nakazato, and M.~Tsukada, ``Limitations and future aspects of communication costs in federated learning: A survey,'' \emph{Sensors}, vol.~23, no.~17, 2023. [Online]. Available: \url{https://www.mdpi.com/1424-8220/23/17/7358}
\BIBentrySTDinterwordspacing

\bibitem{han2020adaptive}
P.~Han, S.~Wang, and K.~K. Leung, ``Adaptive gradient sparsification for efficient federated learning: An online learning approach,'' in \emph{IEEE International Conference on Distributed Computing Systems (ICDCS)}, 2020, pp. 300--310.

\bibitem{aji2017sparse}
A.~Aji and K.~Heafield, ``Sparse communication for distributed gradient descent,'' in \emph{EMNLP 2017: Conference on Empirical Methods in Natural Language Processing}.\hskip 1em plus 0.5em minus 0.4em\relax Association for Computational Linguistics (ACL), 2017, pp. 440--445.

\bibitem{alistarh2018convergence}
D.~Alistarh, T.~Hoefler, M.~Johansson, N.~Konstantinov, S.~Khirirat, and C.~Renggli, ``The convergence of sparsified gradient methods,'' \emph{Advances in Neural Information Processing Systems}, vol.~31, 2018.

\bibitem{shi2019topksparsification}
\BIBentryALTinterwordspacing
S.~Shi, X.~Chu, K.~C. Cheung, and S.~See, ``Understanding top-k sparsification in distributed deep learning,'' 2019. [Online]. Available: \url{https://arxiv.org/abs/1911.08772}
\BIBentrySTDinterwordspacing

\bibitem{wen2017terngrad}
W.~Wen, C.~Xu, F.~Yan, C.~Wu, Y.~Wang, Y.~Chen, and H.~Li, ``Terngrad: ternary gradients to reduce communication in distributed deep learning,'' in \emph{Proceedings of the 31st International Conference on Neural Information Processing Systems}, 2017, p. 1508–1518.

\bibitem{yujun2018deepgradientcompression}
\BIBentryALTinterwordspacing
Y.~Lin, S.~Han, H.~Mao, Y.~Wang, and W.~Dally, ``Deep gradient compression: Reducing the communication bandwidth for distributed training,'' 2018. [Online]. Available: \url{https://openreview.net/pdf?id=SkhQHMW0W}
\BIBentrySTDinterwordspacing

\bibitem{shlezinger2021UVeQFed}
N.~Shlezinger, M.~Chen, Y.~C. Eldar, H.~V. Poor, and S.~Cui, ``Uveqfed: Universal vector quantization for federated learning,'' \emph{IEEE Transactions on Signal Processing}, vol.~69, pp. 500--514, 2021.

\bibitem{lang2025olala}
N.~Lang, M.~Simhi, and N.~Shlezinger, ``{OLAL}a: Online learned adaptive lattice codes for heterogeneous federated learning,'' \emph{arXiv preprint arXiv:2506.20297}, 2025.

\bibitem{tandon2017gradient}
R.~Tandon, Q.~Lei, A.~G. Dimakis, and N.~Karampatziakis, ``Gradient coding: Avoiding stragglers in distributed learning,'' in \emph{International Conference on Machine Learning}.\hskip 1em plus 0.5em minus 0.4em\relax PMLR, 2017, pp. 3368--3376.

\bibitem{chen2016revisiting}
J.~Chen, X.~Pan, R.~Monga, S.~Bengio, and R.~Jozefowicz, ``Revisiting distributed synchronous {SGD},'' \emph{arXiv preprint arXiv:1604.00981}, 2016.

\bibitem{li2019convergence}
X.~Li, K.~Huang, W.~Yang, S.~Wang, and Z.~Zhang, ``On the convergence of fedavg on non-iid data,'' in \emph{International Conference on Learning Representations}, 2019.

\bibitem{xu2023asynchronous}
C.~Xu, Y.~Qu, Y.~Xiang, and L.~Gao, ``Asynchronous federated learning on heterogeneous devices: A survey,'' \emph{Computer Science Review}, vol.~50, p. 100595, 2023.

\bibitem{xie2020asynchronous}
\BIBentryALTinterwordspacing
C.~Xie, S.~Koyejo, and I.~Gupta, ``Asynchronous federated optimization,'' 2020. [Online]. Available: \url{https://arxiv.org/abs/1903.03934}
\BIBentrySTDinterwordspacing

\bibitem{dai2018toward}
W.~Dai, Y.~Zhou, N.~Dong, H.~Zhang, and E.~P. Xing, ``Toward understanding the impact of staleness in distributed machine learning,'' \emph{arXiv preprint arXiv:1810.03264}, 2018.

\bibitem{nguyen2022federated}
J.~Nguyen, K.~Malik, H.~Zhan, A.~Yousefpour, M.~Rabbat, M.~Malek, and D.~Huba, ``Federated learning with buffered asynchronous aggregation,'' in \emph{International conference on artificial intelligence and statistics}.\hskip 1em plus 0.5em minus 0.4em\relax PMLR, 2022, pp. 3581--3607.

\bibitem{pfeiffer2023federated}
K.~Pfeiffer, M.~Rapp, R.~Khalili, and J.~Henkel, ``Federated learning for computationally constrained heterogeneous devices: A survey,'' \emph{ACM Computing Surveys}, vol.~55, no. 14s, pp. 1--27, 2023.

\bibitem{ma2021fedsa}
Q.~Ma, Y.~Xu, H.~Xu, Z.~Jiang, L.~Huang, and H.~Huang, ``Fed{SA}: A semi-asynchronous federated learning mechanism in heterogeneous edge computing,'' \emph{{IEEE} Journal on Selected Areas in Communications}, vol.~39, no.~12, pp. 3654--3672, 2021.

\bibitem{zhang2023semi}
J.~Zhang, W.~Liu, Y.~He, Z.~He, and M.~Guizani, ``Semi-asynchronous model design for federated learning in mobile edge networks,'' \emph{{IEEE} Transactions on Vehicular Technology}, vol.~72, no.~12, pp. 16\,280--16\,292, 2023.

\bibitem{nishio2019client}
T.~Nishio and R.~Yonetani, ``Client selection for federated learning with heterogeneous resources in mobile edge,'' in \emph{ICC 2019-2019 IEEE international conference on communications (ICC)}.\hskip 1em plus 0.5em minus 0.4em\relax IEEE, 2019, pp. 1--7.

\bibitem{zhang2023timelyfl}
T.~Zhang, L.~Gao, S.~Lee, M.~Zhang, and S.~Avestimehr, ``Timelyfl: Heterogeneity-aware asynchronous federated learning with adaptive partial training,'' in \emph{Proceedings of the IEEE/CVF Conference on Computer Vision and Pattern Recognition}, 2023, pp. 5064--5073.

\bibitem{hu2023scheduling}
C.-H. Hu, Z.~Chen, and E.~G. Larsson, ``Scheduling and aggregation design for asynchronous federated learning over wireless networks,'' \emph{{IEEE} Journal on Selected Areas in Communications}, vol.~41, no.~4, pp. 874--886, 2023.

\bibitem{li2020federated}
T.~Li, A.~Sahu, A.~Talwalkar, and V.~Smith, ``Federated optimization in heterogeneous networks,'' \emph{Proceedings of MLSys}, 2020, fedProx, empirical evaluation on heterogeneous clients.

\bibitem{diao2020heterofl}
E.~Diao, J.~Ding, and V.~Tarokh, ``{HeteroFL}: Computation and communication efficient federated learning for heterogeneous clients,'' \emph{arXiv preprint arXiv:2010.01264}, 2020.

\bibitem{lang2024stragglers}
N.~Lang, A.~Cohen, and N.~Shlezinger, ``Stragglers-aware low-latency synchronous federated learning via layer-wise model updates,'' \emph{IEEE Transactions on Communications}, vol.~73, no.~5, pp. 3333--3346, 2025.

\bibitem{bonawitz2019towards}
K.~Bonawitz, H.~Eichner, W.~Grieskamp, D.~Huba, A.~Ingerman, V.~Ivanov, C.~Kiddon, J.~Kone{\v{c}}n{\`y}, S.~Mazzocchi, B.~McMahan \emph{et~al.}, ``Towards federated learning at scale: System design,'' \emph{Proceedings of machine learning and systems}, vol.~1, pp. 374--388, 2019.

\bibitem{deng2012mnist}
L.~Deng, ``The {MNIST} database of handwritten digit images for machine learning research [best of the web],'' \emph{IEEE signal processing magazine}, vol.~29, no.~6, pp. 141--142, 2012.

\bibitem{krizhevsky2009learning}
\BIBentryALTinterwordspacing
A.~Krizhevsky and G.~Hinton, ``Learning multiple layers of features from tiny images,'' 2009. [Online]. Available: \url{https://www.cs.utoronto.ca/~kriz/learning-features-2009-TR.pdf}
\BIBentrySTDinterwordspacing

\bibitem{li2023fedtcr}
K.~Li, H.~Wang, and Q.~Zhang, ``{FedTCR}: Communication-efficient federated learning via taming computing resources,'' \emph{Complex \& Intelligent Systems}, vol.~9, no.~5, pp. 5199--5219, 2023.

\bibitem{park2021few}
Y.~Park, D.-J. Han, D.-Y. Kim, J.~Seo, and J.~Moon, ``Few-round learning for federated learning,'' \emph{Advances in Neural Information Processing Systems}, vol.~34, pp. 28\,612--28\,622, 2021.

\bibitem{ma2023adapbatch}
Z.~Ma, Y.~Xu, H.~Xu, Z.~Meng, L.~Huang, and Y.~Xue, ``Adaptive batch size for federated learning in resource-constrained edge computing,'' \emph{IEEE Transactions on Mobile Computing}, vol.~22, no.~1, pp. 37--53, 2023.

\bibitem{liu2023adacoopt}
W.~Liu, X.~Zhang, J.~Duan, C.~Joe-Wong, Z.~Zhou, and X.~Chen, ``{AdaCoOpt}: Leverage the interplay of batch size and aggregation frequency for federated learning,'' in \emph{IEEE/ACM International Symposium on Quality of Service (IWQoS)}, 2023.

\bibitem{wu2021fast}
H.~Wu and P.~Wang, ``Fast-convergent federated learning with adaptive weighting,'' \emph{IEEE Transactions on Cognitive Communications and Networking}, vol.~7, no.~4, pp. 1078--1088, 2021.

\bibitem{fraboni2023general}
Y.~Fraboni, R.~Vidal, L.~Kameni, and M.~Lorenzi, ``A general theory for federated optimization with asynchronous and heterogeneous clients updates,'' \emph{Journal of Machine Learning Research}, vol.~24, no. 110, pp. 1--43, 2023.

\bibitem{dutta2018slow}
S.~Dutta, G.~Joshi, S.~Ghosh, P.~Dube, and P.~Nagpurkar, ``Slow and stale gradients can win the race: Error-runtime trade-offs in distributed {SGD},'' in \emph{International conference on artificial intelligence and statistics}.\hskip 1em plus 0.5em minus 0.4em\relax PMLR, 2018, pp. 803--812.

\bibitem{shi2020device}
W.~Shi, S.~Zhou, and Z.~Niu, ``Device scheduling with fast convergence for wireless federated learning,'' in \emph{IEEE International Conference on Communications (ICC)}, 2020.

\bibitem{lee2017speeding}
K.~Lee, M.~Lam, R.~Pedarsani, D.~Papailiopoulos, and K.~Ramchandran, ``Speeding up distributed machine learning using codes,'' \emph{IEEE Transactions on Information Theory}, vol.~64, no.~3, pp. 1514--1529, 2017.

\bibitem{jia2024efficient}
J.~Jia, J.~Liu, C.~Zhou, H.~Tian, M.~Dong, and D.~Dou, ``Efficient asynchronous federated learning with sparsification and quantization,'' \emph{Concurrency and Computation: Practice and Experience}, vol.~36, no.~9, p. e8002, 2024.

\bibitem{esfahanizadeh2022stream}
H.~Esfahanizadeh, A.~Cohen, and M.~M{\'e}dard, ``Stream iterative distributed coded computing for learning applications in heterogeneous systems,'' in \emph{IEEE Conference on Computer Communications}, 2022, pp. 230--239.

\bibitem{zhao2018federated}
Y.~Zhao, M.~Li, L.~Lai, N.~Suda, D.~Civin, and V.~Chandra, ``Federated learning with non-iid data,'' \emph{arXiv preprint arXiv:1806.00582}, 2018.

\bibitem{dlmf}
{NIST Digital Library of Mathematical Functions}, ``{Chapter 8. Gamma and Related Functions},'' \url{https://dlmf.nist.gov/8.2}, 2023, release 1.1.9 of 2023-03-15.

\bibitem{yuan2000review}
Y.-x. Yuan, ``A review of trust region algorithms for optimization,'' \emph{ICM99: Proceedings of the Fourth International Congress on Industrial and Applied Mathematics}, 09 1999.

\bibitem{hsu2019measuring}
\BIBentryALTinterwordspacing
T.-M.~H. Hsu, H.~Qi, and M.~Brown, ``Measuring the effects of non-identical data distribution for federated visual classification,'' 2019. [Online]. Available: \url{https://arxiv.org/abs/1909.06335}
\BIBentrySTDinterwordspacing

\bibitem{simonyan2015very}
\BIBentryALTinterwordspacing
K.~Simonyan and A.~Zisserman, ``Very deep convolutional networks for large-scale image recognition,'' \emph{International Conference on Learning Representations (ICLR)}, 2015. [Online]. Available: \url{https://arxiv.org/abs/1409.1556}
\BIBentrySTDinterwordspacing

\bibitem{loshchilov2019decoupled}
I.~Loshchilov and F.~Hutter, ``Decoupled weight decay regularization,'' in \emph{International Conference on Learning Representations (ICLR)}, 2019, empirical study of weight decay and optimizer behavior.

\bibitem{goodfellow2016deep}
I.~Goodfellow, Y.~Bengio, and A.~Courville, \emph{Deep Learning}.\hskip 1em plus 0.5em minus 0.4em\relax MIT Press, 2016, \url{http://www.deeplearningbook.org}.

\bibitem{karimireddy2020scaffold}
S.~P. e.~a. Karimireddy, ``Scaffold: Stochastic controlled averaging for federated learning,'' in \emph{ICML}, 2020.

\bibitem{ross2014introduction}
S.~M. Ross, \emph{Introduction to probability models}.\hskip 1em plus 0.5em minus 0.4em\relax Academic press, 2014.

\end{thebibliography}

\end{document}